\documentclass{article}

\usepackage[utf8]{inputenc} 
\usepackage[T1]{fontenc}    
\usepackage{hyperref}       
\usepackage{url}            
\usepackage[numbers]{natbib}

\usepackage{amsmath,amssymb,amsfonts}
\usepackage{graphicx}
\usepackage{textcomp}

\usepackage{color}
\usepackage{colortbl}
\usepackage{xcolor} 

\definecolor{Gray}{gray}{1}

\usepackage{multirow}
\usepackage{makecell}
\usepackage{pifont}
\usepackage{booktabs}
\usepackage{bm}
\usepackage{url}
\usepackage{booktabs}
\usepackage{algorithm}
\usepackage{algpseudocode}
\usepackage{calc}

\newcommand{\cellimage}[1]{%
    \raisebox{-0.25\height}{\includegraphics[height=\heightof{\strut}]{#1}}%
}

\newcommand{\tabimg}{\includegraphics[height=\heightof{\strut}]{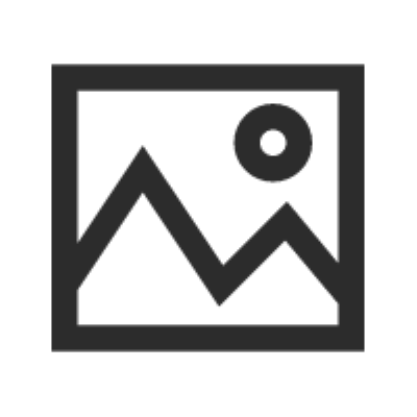}}

\newcommand{\tabvideo}{\includegraphics[height=\heightof{\strut}]{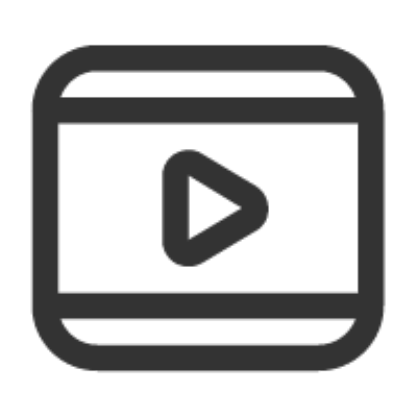}}

\definecolor{myRed}{RGB}{195,10,10}
\definecolor{myGreen}{RGB}{55,149,73}

\newcommand{\cmark}{\textcolor{myGreen}{\ding{51}}}
\newcommand{\xmark}{\textcolor{myRed}{\ding{55}}}

\definecolor{myRed2}{RGB}{255, 0, 0}

\begin{document}

\title{Reasoning Segmentation for Images and Videos: A Survey}
\author{
  Yiqing Shen\textsuperscript{1}, Chenjia Li\textsuperscript{1}, Fei Xiong\textsuperscript{2}, \\
  Jeong-O Jeong\textsuperscript{2}, Tianpeng Wang\textsuperscript{2}, Michael Latman\textsuperscript{2}, \\Mathias Unberath\textsuperscript{1} \\[2ex]
  \textsuperscript{1} Johns Hopkins University \\
  \textsuperscript{2} Amazon Web Services
}

\date{}
\maketitle

\begin{abstract}
Reasoning Segmentation (RS) aims to delineate objects based on implicit text queries, the interpretation of which requires reasoning and knowledge integration. 
Unlike the traditional formulation of segmentation problems that relies on fixed semantic categories or explicit prompting, RS bridges the gap between visual perception and human-like reasoning capabilities, facilitating more intuitive human-AI interaction through natural language.
Our work presents the first comprehensive survey of RS for image and video processing, examining 26 state-of-the-art methods together with a review of the corresponding evaluation metrics, as well as 29 datasets and benchmarks.
We also explore existing applications of RS across diverse domains and identify their potential extensions.
Finally, we identify current research gaps and highlight promising future directions. 
\end{abstract}



\section{Introduction}

By identifying and delineating objects or regions of interest in images and videos, segmentation is among the most fundamental computer vision tasks that enable machines to parse visual environments at different levels of abstraction \cite{minaee2021image}.
The field of segmentation has evolved through several distinct paradigms.
Semantic segmentation, the most basic form, assigns predefined semantic categories to each pixel in an image, enabling scene parsing without distinguishing between instances of the same class \cite{hao2020brief}. 
Instance segmentation advances this approach by differentiating individual objects within the same semantic category, such as distinguishing multiple vehicles in a street scene \cite{hafiz2020survey}. 
Panoptic segmentation further unifies these approaches by handling both countable foreground objects in an instance segmentation manner and uncountable background elements semantically, offering a more complete scene decomposition \cite{elharrouss2021panoptic}.
In addition to the aforementioned distinctions, segmentation paradigms can be categorized by the level of automation or user involvement they allow and/or require.

Interactive segmentation methods have become an important paradigm, particularly in domains where high precision is required, but fully automatic solutions fall short. 
Traditional interactive segmentation approaches such as GrowCut \cite{vezhnevets2005growcut} and Graph Cuts facilitate user-guided segmentation through pictorial input such as scribbles, box bounding, or seed points. 
These methods are especially valuable in medical imaging, where expert knowledge is important for the accurate delineation of anatomical structures \cite{amrehn2017ui,shen2025performance}.
Recent advances in deep learning have transformed interactive segmentation. 
Approaches such as UI-Net \cite{amrehn2017ui} incorporate user feedback directly into neural network architectures, enabling iterative refinement through a cooperative human-machine process. 
In this framework, users provide a visual prompt in the form of scribbles or seed points, which the network interprets to progressively improve segmentation results. 
The Segment Anything Model (SAM) \cite{sam1, sam2} represents another major advancement in this direction, designed to segment any object based on user prompts such as points or boxes, demonstrating remarkable zero-shot generalization capabilities across various visual inputs.

Moving beyond visual prompting, referring segmentation enables natural language as an interface for specifying targets for segmentation through direct descriptive text queries (\textit{e}.\textit{g}.,``\textit{the red cup}'', ``\textit{the person wearing a blue shirt}'') \cite{refcoco,refcocog,giou,yang2022lavt}. 
This approach bridges vision and language, enabling more intuitive human-machine interaction for segmentation tasks. 
Despite the increased flexibility afforded by language-based prompting, the need for explicitly defining object categories or directly observable visual attributes limits the practical applicability of these methods in complex real-world scenarios where human instructions are often implicit and require reasoning.

\begin{figure}[t!]
    \centering
    \includegraphics[width=1.0\linewidth]{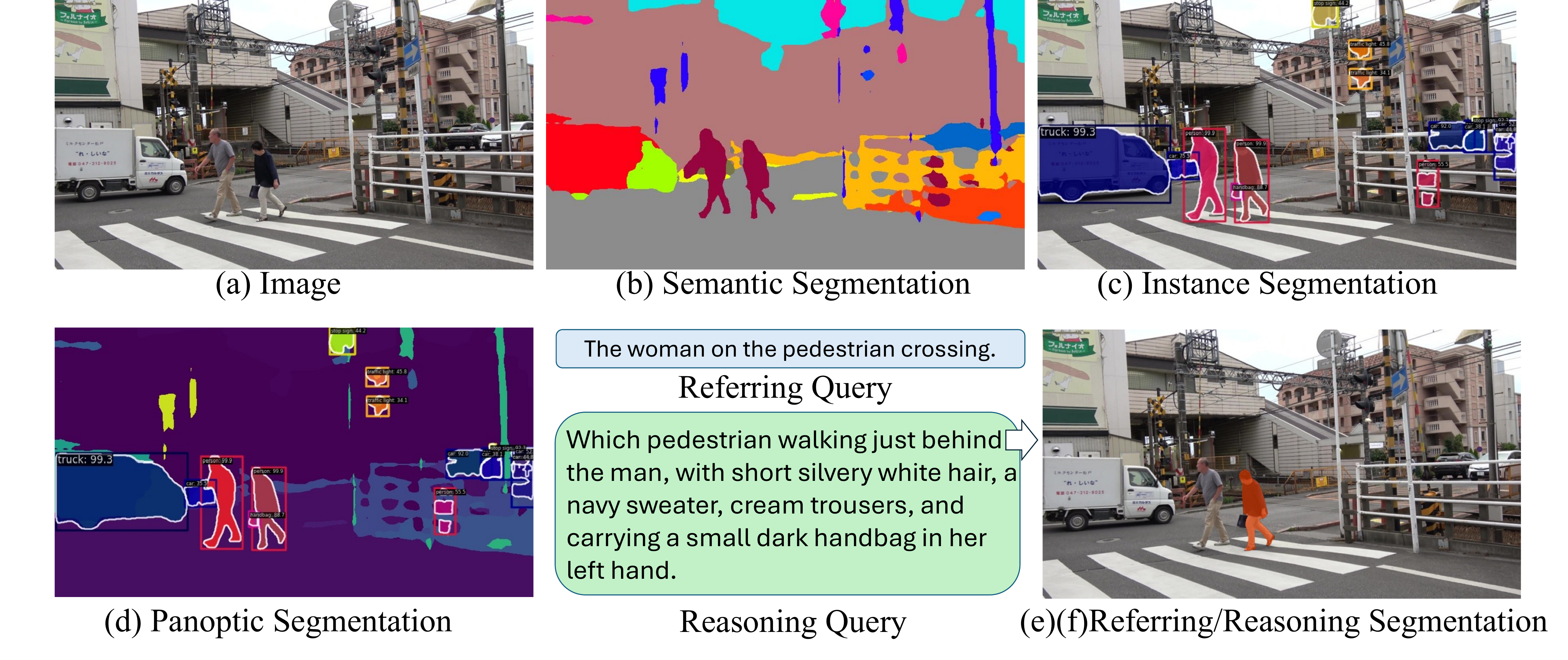}
    \caption{
    Illustration of various segmentation tasks on a street-crossing scene. 
    (a) Original image of the scene. 
    (b) Semantic segmentation where each pixel is assigned to a predefined category (\textit{e}.\textit{g}., person, road, vehicle) without distinguishing individual instances. 
    (c) Instance segmentation, which differentiates individual objects within the same semantic category (\textit{e}.\textit{g}., multiple pedestrians). 
    (d) Panoptic segmentation, combining both countable foreground objects and uncountable background elements. 
    The upper text bubble shows a ``referring query'' with a direct descriptive expression (``\textit{The woman on the pedestrian crossing}''), while the lower bubble presents a ``reasoning query'' requiring multi-step inference (``\textit{Which pedestrian walking just behind the man, with short silvery white hair, a navy sweater, cream trousers, and carrying a small dark handbag in her left hand}''). 
Panel (e)/(f) shows the resulting segmentation masks for both the referring and reasoning queries, highlighting how reasoning segmentation can handle more complex, implicit descriptions beyond directly observable attributes. 
    }\label{fig:segcat}
\end{figure}

To address this limitation, reasoning segmentation (RS) has been proposed \cite{lisa}. 
A comparison of all these segmentation tasks is shown in Fig.~\ref{fig:segcat}.
Specifically, RS requires segmentation models to understand implicit relationships, whether spatial, temporal, functional, or causal, to segment the corresponding object in a manner that better mirrors human cognitive processes.
Unlike direct text queries in referring segmentation, RS aims to handle implicit queries that demand multi-step reasoning and integration of prior knowledge. 
For example, when faced with questions like ``\textit{Which object in this kitchen can keep food fresh for a longer time?}'' or ``\textit{What would be most dangerous if a child touches it?}'', RS models must engage in reasoning beyond basic visual recognition. 
Of course, RS is only the first step towards even more general (embodied) AI assistants that will be enabled by RS-based methods, for example appropriate responses to queries like ``\textit{Please stow away the groceries.}'' or ``\textit{Secure the are for a visit of friends with children.}'' for the two RS prompts above, respectively.  
The reasoning process can involve multiple stages, such as:
(1) scene understanding to identify all relevant objects and their relationships, 
(2) interpretation of abstract concepts present in the implicit text query (\textit{e}.\textit{g}., ``\textit{freshness}'', ``\textit{danger}''), 
(3) application of world knowledge combined with logical, spatial, temporal, or other kinds of reasoning to connect visual elements with query requirements,
and (4) precise mask generation for the identified targets. 
Based on visual input modalities, we categorize RS into two main domains in this paper: (1) image RS for static images, (2) video RS for video sequences, as shown in Table~\ref{table:rs_feature_comparison}.
The emergence of RS coincides with advances in large language models (LLMs) and multimodal large language models (MLLMs). 
These large models have demonstrated impressive capabilities in understanding complex instructions, performing multi-step reasoning, and generating coherent responses across various domains. 
Early attempts, such as LISA \cite{lisa} and VISA \cite{visa} have pioneered the integration of MLLMs with segmentation capabilities by reasoning about implicit queries and producing the corresponding pixel-level segmentation masks.

\begin{figure}[t!]
    \centering
    \includegraphics[width=1.0\linewidth]{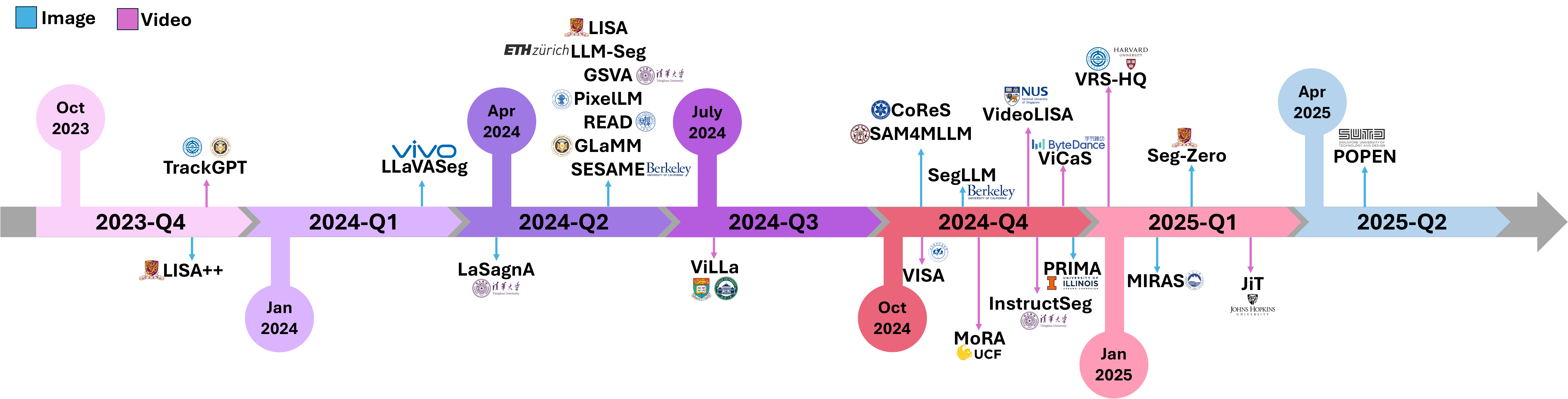}
    \caption{
Timeline of RS methods from October 2023 to April 2025, illustrating its rapid evolution. 
Blue and purple labels distinguish between image-based and video-based RS approaches, respectively. 
The timeline showcases the sequential development of these methods, beginning with LISA as the pioneering work, followed by numerous innovations expanding the capabilities of RS across both modalities.
    }\label{fig:timeline}
\end{figure}

Several surveys have previously explored different aspects of segmentation and visual understanding. 
For example, Minaee \textit{et al.} \cite{minaee2021image} provided an overview of deep learning methods for image segmentation, Hafiz \textit{et al.} \cite{hafiz2020survey} focused specifically on instance segmentation approaches, while Khan \textit{et al.} \cite{khan2022transformers} concentrate on transformer-based segmentation model architectures. 
More recently, Zhang et al. \cite{zhang2024vision} surveyed vision language models for various visual recognition tasks, including segmentation, and Chen \textit{et al.} \cite{chen2023vlp} explored vision language pretraining in both the image text and video text domains. 
However, no existing survey has specifically focused on RS, which represents a shift in how ML models understand and interact with visual content.
This survey aims to fill this gap by providing a comprehensive review of RS in two visual modalities (image and video). 
As illustrated in Fig.~\ref{fig:timeline}, RS has experienced rapid development since its introduction in late 2023, with 26 distinct methods proposed across image and video modalities in less than two years, further emphasizing the need for a comprehensive survey of this emerging field.

The major contributions of this survey are three-fold.
First, this is the first survey paper on RS, reviewing 17 image RS methods and 9 video RS methods developed to date.
Specifically, we provide a categorization and analysis of these RS methods for both images and videos, covering architectural designs and reasoning mechanisms.
Second, we provide an in-depth examination of evaluation metrics, datasets, and benchmarks specifically designed for RS, highlighting their characteristics and challenges across 19 image RS datasets and 10 video RS datasets.
Third, we discuss current challenges, limitations, and future research directions in RS.

\begin{table*}[t!]
\caption{
Comparison of RS methods across multiple dimensions. 
The comparison spans five dimensions: 
(1) ``Modality'' evaluates each method's ability to handle different input types with icons for image (\tabimg) and video (\tabvideo). A single check mark (\cmark) indicates basic support, while a double check mark (\cmark\cmark) indicates support for multiple instances. 
(2) ``Segmentation Capabilities'' assesses semantic segmentation functionality (where \cmark\cmark indicates multi-category support), instance-level segmentation, and empty-target rejection.
(3) ``Architectural Features'' examines implementation including conversational abilities, fine-tuning requirements, reasoning approach, and architectural design. 
(4) ``Processing Approach'' distinguishes between online methods (processing frames sequentially) and offline methods (processing all frames at once), and includes information about temporal consistency mechanisms.
(5) ``Implementation Details'' covers external model dependencies (e.g., SAM, XMem), specialized tokens used, and training methodology (E2E: End-to-End, RL: Reinforcement Learning, SFT: Supervised Fine-Tuning, ZS: Zero-Shot).
}
\label{table:rs_feature_comparison}
\centering
\resizebox{\linewidth}{!}{
\begin{tabular}{l|l|cc|ccc|cccc|cc|ccc}
\toprule
\multirow{2}{*}{Methods} & \multirow{2}{*}{Venues} & \multicolumn{2}{c|}{Modalities} & \multicolumn{3}{c|}{Segmentation Capabilities} & \multicolumn{4}{c|}{Architectural Features} & \multicolumn{2}{c|}{Processing Approach} & \multicolumn{3}{c}{Implementation Details} \\
\cline{3-16}
& & \tabimg & \tabvideo & Semantic & Instance & \makecell[c]{Reject\\Empty} & \makecell[c]{Multi-Round\\Conversation} & \makecell[c]{Fine-Tuning\\Free of MLLM} & \makecell[c]{CoT\\Reasoning} & \makecell[c]{Decoupled\\Architecture} & \makecell[c]{Online/\\Offline} & \makecell[c]{Temporal\\Consistency} & \makecell[c]{External\\Dependencies} & \makecell[c]{Specialized\\Tokens} & \makecell[c]{Training\\Method} \\
\hline
\rowcolor{gray!5}
LISA \cite{lisa} & CVPR-24 & \cmark & \xmark & \cmark & \xmark & \xmark & \xmark & \xmark & \xmark & \xmark & - & - & SAM & <SEG> & SFT \\
LISA++ \cite{lisa++} & arXiv-23 & \cmark & \xmark & \cmark\cmark & \cmark & \xmark & \cmark & \xmark & \xmark & \xmark & - & - & SAM & <SEG> & SFT \\
\rowcolor{gray!5}
LLM-Seg \cite{llmseg} & CVPRw-24 & \cmark & \xmark & \cmark & \xmark & \xmark & \xmark & \xmark & \xmark & \cmark & - & - & SAM, DINOv2 & <SEG> & SFT \\
GSVA \cite{gsva} & CVPR-24 & \cmark & \xmark & \cmark & \cmark & \cmark & \xmark & \xmark & \xmark & \xmark & - & - & SAM & <SEG>, <REJ> & SFT \\
\rowcolor{gray!5}
LaSagnA \cite{lasagna} & arXiv-24 & \cmark & \xmark & \cmark\cmark & \xmark & \cmark & \xmark & \xmark & \xmark & \xmark & - & - & SAM & <SEG>, <NEG> & SFT \\
CoReS \cite{cores} & ECCV-24 & \cmark & \xmark & \cmark & \xmark & \xmark & \xmark & \xmark & \cmark & \cmark & - & - & SAM & <SEG> & SFT \\
\rowcolor{gray!5}
SegLLM \cite{segllm} & arXiv-24 & \cmark & \xmark & \cmark & \xmark & \xmark & \cmark & \xmark & \xmark & \xmark & - & - & CLIP, SAM & <SEG>, <REF> & SFT \\
PRIMA \cite{prima} & arXiv-24 & \cmark\cmark & \xmark & \cmark \cmark & \cmark & \xmark & \xmark & \xmark & \xmark & \xmark & - & - & DINOv2, EVA-CLIP, SAM & <SEG> & SFT \\
\rowcolor{gray!5}
LLaVASeg \cite{llavaseg} & arXiv-24 & \cmark & \xmark & \cmark & \xmark & \xmark & \xmark & \cmark & \cmark & \cmark & - & - & SAM & - & SFT \\
SAM4MLLM \cite{sam4mllm} & ECCV-24 & \cmark & \xmark & \cmark & \xmark & \xmark & \xmark & \xmark & \xmark & \cmark & - & - & SAM & - & SFT \\
\rowcolor{gray!5}
PixelLM \cite{pixellm} & CVPR-24 & \cmark & \xmark & \cmark\cmark & \xmark & \xmark & \xmark & \xmark & \xmark & \xmark & - & - & CLIP & <SEG> & SFT \\
READ \cite{read} & CVPR-24 & \cmark & \xmark & \cmark & \xmark & \xmark & \xmark & \xmark & \xmark & \cmark & - & - & SAM & <SEG> & SFT \\
\rowcolor{gray!5}
GLaMM \cite{glamm} & CVPR-24 & \cmark & \xmark & \cmark\cmark & \cmark & \xmark & \cmark & \xmark & \xmark & \xmark & - & - & SAM & <SEG> & E2E \\
SESAME \cite{sesame} & CVPR-24 & \cmark & \xmark & \cmark & \cmark & \cmark & \xmark & \xmark & \xmark & \xmark & - & - & SAM & <SEG> & SFT \\
\rowcolor{gray!5}
Seg-Zero \cite{segzero} & arXiv-25 & \cmark & \xmark & \cmark & \xmark & \xmark & \xmark & \xmark & \cmark & \cmark & - & - & SAM & - & RL \\
MIRAS \cite{miras} & arXiv-25 & \cmark & \xmark & \cmark & \cmark & \xmark & \cmark & \xmark & \cmark & \xmark & - & - & - & <SEG> & SFT \\
\rowcolor{gray!5}
POPEN \cite{popen} & CVPR-25 & \cmark & \xmark & \cmark\cmark & \cmark & \xmark & \xmark & \xmark & \xmark & \xmark & - & - & CLIP & <SEG> & SFT+RL \\
\hline
\rowcolor{gray!5}
TrackGPT \cite{trackgpt} & arXiv-23 & \cmark & \cmark & \cmark & \xmark & \xmark & \xmark & \xmark & \xmark & \xmark & Online & \cmark & SAM & <TK>, <PO> & SFT \\
VISA \cite{visa} & ECCV-24 & \cmark & \cmark & \cmark & \xmark & \xmark & \xmark & \xmark & \xmark & \xmark & Offline & \cmark & SAM, LLaMA-VID, XMem & <SEG> & SFT \\
\rowcolor{gray!5}
ViLLa \cite{villa}  & arXiv-24  & \xmark & \cmark & \cmark\cmark & \cmark & \xmark & \xmark & \xmark & \xmark & \xmark & Online & \cmark & - & Multiple & E2E \\
VideoLISA \cite{videolisa} & NeurIPS-24 & \cmark & \cmark & \cmark & \xmark & \xmark & \xmark & \xmark & \xmark & \xmark & Offline & \cmark & SAM, XMem++ & <TRK> & E2E \\
\rowcolor{gray!5}
InstructSeg \cite{instructseg} & arXiv-24 & \cmark & \cmark & \cmark\cmark & \cmark & \xmark & \xmark & \xmark & \xmark & \xmark & Offline & \cmark & CLIP & Multiple & SFT \\
VRS-HQ \cite{vrshq} & CVPR-25 & \cmark & \cmark & \cmark & \xmark & \xmark & \xmark & \xmark & \xmark & \xmark & Offline & \cmark\cmark & SAM2 & <SEG>, <TAK> & E2E \\
\rowcolor{gray!5}
MoRA \cite{mora} & CVPR-25 & \cmark & \cmark & \cmark & \xmark & \xmark & \xmark & \xmark & \xmark & \xmark & Online & \cmark & CLIP & <SEG>, <LOC> & SFT \\
ViCaS \cite{vicas} & CVPR-25 & \xmark & \cmark & \cmark & \cmark & \xmark & \xmark & \xmark & \xmark & \xmark & Offline & \cmark & - & <SEG> & E2E \\
\rowcolor{gray!5}
JiT \cite{jitdt} & arXiv-25 & \cmark & \cmark & \cmark & \cmark & \cmark & \xmark & \cmark & \cmark & \cmark & Online & \cmark & SAM, DepthAnything & - & ZS \\
\bottomrule
\end{tabular}
}
\end{table*}

\section{Image Reasoning Segmentation}

\subsection{Task Definition}

Image RS focuses on generating segmentation masks at the pixel level based on implicit text queries \cite{lisa}.
Therefore, it requires visual understanding of the image, together with reasoning capabilities to identify regions that satisfy abstract or indirect requirements specified in natural language.
The image RS task can be formally defined as follows:
Given an input image $\mathbf{X}_\text{img} \in \mathbb{R}^{H \times W \times 3}$, where $H$ and $W$ denote the height and width dimensions in pixels, and an implicit text query $Q$, the goal is to produce a binary segmentation mask $\mathbf{M} \in \{0,1\}^{H \times W}$ that identifies the target regions that meet the reasoning requirements specified in $Q$.
The output mask $M$ is defined for each pixel position $(i,j)$ as:
\begin{equation}
\mathbf{M}(i,j) = \begin{cases}
1 & \text{if pixel } (i,j) \text{ belongs to the reasoned target described by }$Q$ \\
0 & \text{otherwise}
\end{cases}.
\label{eq:image_rs_mask}
\end{equation}
Then, image RS can be formulated as:
\begin{equation}
\mathbf{M} = \varphi_{\theta}(\mathbf{X}_\text{img},Q),
\end{equation}
where $\varphi_{\theta}(\cdot)$ denotes the RS model with parameters $\theta$.

\subsection{Methods}

Multimodal large language models (MLLMs) have demonstrated their ability to understand both user intentions and visual content through zero-shot generalization, making them promising candidates for handling implicit text queries required in RS.
However, while MLLMs excel at text generation, they traditionally lack the ability to produce segmentation masks.
To address this limitation, previous work has focused on architectural changes and fine-tuning MLLMs to enable them to generate masks from the end to the end for RS \cite{lisa,lisa++,gsva,lasagna,segllm,prima}; or use them to trigger a separate segmentation network \cite{llmseg,cores,llavaseg,sam4mllm}.

\begin{figure}[t!]
    \centering
    \includegraphics[width=1.0\linewidth]{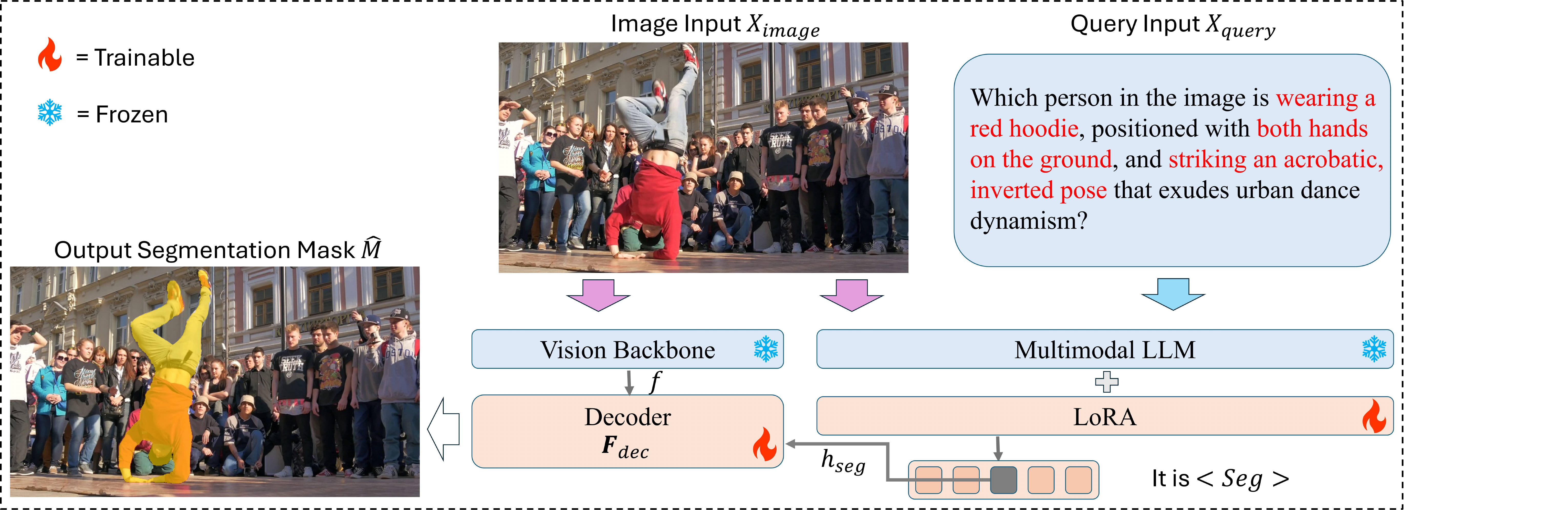}
    \caption{
    Overview of the LISA \cite{lisa}, which adopts an ``embedding-as-mask'' for RS. 
    }\label{fig:lisa}
\end{figure}

\paragraph{LISA}
The first attempt at image RS is the language instructed segmentation assistant (LISA) \cite{lisa}, which introduces an ``\textit{embedding-as-mask}'' paradigm that formulates RS masks as continuous embedding vectors to be decoded from MLLM, as depicted in Fig.~\ref{fig:lisa}.
Specifically, LISA employs MLLMs such as LLaVA \cite{llava} to embedding to generate RS masks while preserving their reasoning abilities in processing implicit text queries. 
LISA comprises three components, namely (1) a vision encoder (\textit{e}.\textit{g}., the encoder of SAM \cite{sam1}) for extracting image features, (2) a MLLM for processing images and implicit text queries jointly to identify the objects, and (3) a mask decoder to transform continuous embedding vectors generated by MLLM into RS masks.
For the MLLM component, it expands the MLLM's vocabulary with a \texttt{<SEG>} token that signals segmentation mask output, therefore, the output of MLLM looks like ``\textit{It is \texttt{<SEG>}}''.
When MLLM generates this token, its corresponding hidden embedding will then be transformed into an RS mask through the mask decoder. 
To maintain efficiency and preserve pre-trained knowledge in MLLM while adapting to RS task, LISA employs low-rank adaptation (LoRA) \cite{lora} to fine-tune the MLLM by updating the MLLM's projection layers and decoder parameters while keeping the vision encoder frozen. 
In the fine-tuning process, the MLLM learns to generate the text response together with the mask prediction.

\paragraph{LISA++}
LISA primarily focuses on semantic segmentation by assigning \texttt{<SEG>} token to each semantic category, which therefore fails to distinguish different instances of the same semantic category.
Moreover, the response format of LISA heavily relies on predefined templates during the training stage, which therefore miss the response diversity and can hinder its practical applicability to provide informative text responses.
To this end, LISA++ \cite{lisa++} extends LISA from semantic RS to an instance segmentation manner by making two enhancements to LISA while maintaining its architecture and ``\textit{embedding-as-mask}'' paradigm. 
First, LISA++ incorporates instance segmentation capabilities by adopting bipartite matching during the fine-tuning stage, similar to MaskFormer and DETR \cite{maskformer,detr}, allowing LISA++ to differentiate between instances of the same category by assigning \texttt{<SEG>} token to each instance.
Second, LISA++ introduces the concept of segmentation in dialogue, which enables more natural and flexible text responses conversationally in addition to the RS mask results. 
Consequently, rather than following the fixed format of ``\textit{Sure, it is \texttt{<SEG>}}'' by LISA, LISA++ can seamlessly incorporate segmentation \texttt{<SEG>} tokens at contextually appropriate points in its responses.

\paragraph{LLM-Seg}
Unlike LISA's end-to-end ``\textit{embedding-as-mask}'' paradigm \cite{lisa}, LLM-Seg \cite{llmseg} decouples the implicit text query understanding from segmentation mask generation. 
Similarly to LISA++ \cite{lisa++}, LLM-Seg also focuses on multi-instance RS scenario.
LLM-Seg comprises four components, namely (1) a fine-tuned MLLM (\textit{i}.\textit{e}., LLaVA \cite{llava}) for image and query understanding, 
(2) a frozen SAM \cite{sam1} for generating multiple mask candidates, 
(3) a frozen DINOv2 \cite{dinov2} for extracting visual features, 
and (4) a mask selection module to aggregate LLava's output and DINOv2's feature to pick the appreciate mask candidates from SAM. 
Firstly, SAM generates a series of mask candidates in the segment everything mode, where the image is prompted with a grid of $32\times32$ points uniformly. 
These masks are then refined by mask-pooling DINOv2 features to generate the corresponding mask embeddings. 
Afterward, LLaVA processes the image and the implicit text query to generate the \texttt{<SEG>} token, which is then fused with the mask embeddings to predict the selection scores through the mask selection module.
This selection score measures the alignment of each mask to the \texttt{<SEG>} token to enable multi-instance RS.
Although LLM-Seg decouples reasoning from RS mask generation, it still needs fine-tuning of MLLM to extend the vocabulary with \texttt{<SEG>} token; and training the mask selection module from scratch.

\paragraph{GSVA}
LISA \cite{lisa}, LISA++ \cite{lisa++}, LLM-Seg \cite{llmseg} all fail to identify and handle cases where objects described in the implicit text query are absent from the image, which can lead to hallucination.
To address this gap, the Generalized Segmentation Vision Assistant (GSVA) \cite{gsva} adds a \texttt{<REJ>} token to the LISA's vocabulary to handle cases where the object described in the implicit text query is absent from the image. 
Specifically, MLLM will generate the \texttt{<REJ>} token instead of always generating a \texttt{<SEG>} token that can produce an incorrect segmentation mask.
It therefore liberates the segmentation decoder from attempting to identify non-existent targets and, therefore, can handle empty-target cases.
Importantly, the design of \texttt{<REJ>} token in GSVA can also be integrated into other variants of LISA.

\paragraph{LaSagnA}
LaSagnA \cite{lasagna} aims to handle complex implicit text queries by addressing two main limitations of LISA \cite{lisa}. 
First, compared to LISA, which requires multiple separate queries to process objects of different semantic categories, LaSagnA enables simultaneous processing of multiple semantic objects within a single implicit text query. 
Second, similar to GSVA \cite{gsva}, LaSagnA can also handle cases where the queried objects in the implicit text query are absent from the image through a negative token \texttt{<NEG>}.
By reformulating semantic segmentation in the format of RS where the implicit text query describes all the semantic categories that need to be segmented according to a predefined template, LaSagnA can incorporate multi-semantic segmentation capabilities into the LISA \cite{lisa} by assigning \texttt{<SEG>} token to each semantic category that needs to be segmented.
%
%
Additionally, LaSagnA employs a flexible sequence format that allows more natural handling of complex queries similar to LISA++ \cite{lisa++}, moving beyond the rigid template-based responses in LISA \cite{lisa}.

\paragraph{CoReS}
Chains of Reasoning and Segmenting (CoReS) \cite{cores} approaches a complex implicit text query that requires multi-step reasoning by incorporating chain-of-thoughts (CoT) reasoning \cite{cot} into RS task.
Specifically, it aims to address the challenges of semantic similarity between multiple instances and the limited localization capabilities of MLLMs in RS.
Specifically, CoReS disentangles reasoning with segmentation by a dual-chain architecture, where one chain-of-reasoning aims at semantic-level understanding of text query, and another chain-of-segmenting executes the corresponding segmentation through top-down object localization and searching. 
Similarly to other disentangling methods such as LLM-Seg \cite{llmseg}, CoReS also requires MLLM fine-tuning to organize MLLM's output in a chain-like manner, which can progressively refine the understanding from a broad scene context to specific object details. 
In addition, CoReS also utilizes in-context examples in both the inference and fine-tuning stages.

\paragraph{SegLLM}
SegLLM \cite{segllm} aims to expand the interaction with RS models from a single-round to a multi-round conversational manner.
To maintain the memory of both visual and textual output from previous rounds, SegLLM leverages a mask-aware MLLM that can integrate previous RS results into its input, enabling reasoning about previously segmented objects in a chat-like manner.
SegLLM introduces two architectural modifications to LISA \cite{lisa}: (1) a mask encoder that reintegrates all output masks into the MLLM's input stream and (2) a mask-aware decoder that conditions the mask decoder on both visual and textual conversational history.
Specifically, for each previously generated RS mask, the mask-encoding module computes two types of embedding, namely a semantic mask embedding capturing the masked object's features and a bounding box embedding encoding its spatial location, as a complement of the text and image embedding to the LLM.
SegLLM expands the vocabulary of LISA \cite{lisa} by introducing a \texttt{<REF>} token which encodes information on reference masks from previous rounds.

\paragraph{PRIMA}
PRIMA \cite{prima} expands the scope of RS from single image to multi-image, allowing fine-grained comparison and contextual understanding across multiple images at the pixel level. 
Rather than using a single encoder to embed image (\textit{e}.\textit{g}. encoder of SAM) \cite{lisa,lisa++,gsva}, PRIMA fuses the semantic features of DINOv2 \cite{dinov2} with global features of EVA-CLIP \cite{evaclip} by cross-attention of Q-Former \cite{qformer} to allow precise visual grounding across images while reducing computational demands.
Furthermore, it augments the LISA \cite{lisa} by appending an image identifier associated with \texttt{<SEG>} token to distinguish different images.

\paragraph{LLaVASeg}
LISA \cite{lisa} and its variants usually fine-tune MLLM to expand their vocabularies with \texttt{<SEG>} and other tokens for RS, but at the cost of degrading its original dialogue capability due to forgetting. 
To address this issue, similar to CoReS \cite{cores} and LLM-Seg \cite{llmseg}, LLaVASeg \cite{llavaseg} also disentangles the reasoning and segmentation stages to free the LLaVA from fine-tuning.
To be more specific, it prompts LLaVA in a three-stage CoT manner \cite{cot} to generate explicit descriptions of the target objects, namely the abstract target name and the corresponding visual attributes, which will later be used to prompt a segmentation model to perform the referring segmentation.
Firstly, LLaVA \cite{llava} is prompted to reason about the target region by generating explanations of the reasoning step.
Then, LLaVA extracts the specific target name from the explanations. 
Finally, LLaVA generates the corresponding visual attributes, such as color, shape, and relative position with respect to the target object.
These attributes are then used to initiate a segmentation model trained to generate the final RS masks.
The major idea of LLaVASeg is to convert the RS to a referring segmentation with the MLLM, representing the first attempt at free of MLLM-tuning.

\paragraph{SAM4MLLM}
%
%
Unlike previous RS approaches such as LISA \cite{lisa} and variants \cite{lisa++,llmseg,gsva,lasagna,segllm,prima} that require architectural modifications and specialized tokens (\texttt{<SEG>}) which not only introduce architectural complexity but also complicate the extension of the model for additional tasks, SAM4MLLM \cite{sam4mllm} extends LLaVASeg \cite{llavaseg} by using the semantic awareness of MLLM to prompt SAM \cite{sam1,sam2} for RS.
Specifically, SAM4MLLM proposes two strategies to prompt SAM with MLLM, namely prompt-point generation (PPG) and proactive query of prompt-points (PQPP). 
PPG directly generates prompts for SAM by predicting both a bounding box and point coordinates with a fine-tuned MLLM, while PQPP is more interactive by first predicting a bounding box and then proactively querying whether sampled points lie inside the target region. 
To maintain efficiency and preserve pre-trained knowledge of MLLM, SAM4MLLM utilizes LoRA \cite{lora} for fine-tuning, primarily updating the MLLM's projection layers while keeping the vision encoder frozen.

\paragraph{PixelLM}
LISA \cite{lisa} relies on external pre-trained segmentation models such as SAM \cite{sam1} and is limited to single-target image RS scenarios. PixelLM \cite{pixellm} introduces an efficient end-to-end architecture capable of handling multiple targets of different semantic categories simultaneously, similar to LaSagnA \cite{lasagna}. 
At its core, PixelLM employs a lightweight pixel decoder coupled with a segmentation codebook, enabling direct mask generation from the MLLM without requiring additional segmentation models. 
The segmentation codebook consists of learnable tokens that encode contextual information at different visual granularities or scales. 
At the same time, the pixel decoder transforms these token embeddings alongside image features into segmentation masks based on implicit text queries. 
To enhance multi-target segmentation performance, PixelLM incorporates a token fusion mechanism that combines multiple tokens within each scale group to capture richer semantic information and a target refinement loss that helps differentiate between various targets by focusing on regions where multiple targets might overlap. 
The architecture is trained end-to-end using a combination of auto-regressive cross-entropy loss for text generation, DICE loss for mask generation, and target refinement loss. 

\paragraph{READ} 
Based on observations that the \texttt{<SEG>} token in MLLM primarily functions to learn the semantic correspondences between the text and image regions, READ \cite{read} introduces a framework that leverages this insight to enhance the performance of RS in images by better prompting the decoder region to segment. 
Formally, READ consists of three main components, namely an LLaVA encoder to embed the image and the implicit text query concurrently, a similarity-as-points (SasP) block to generate point prompts, and an SAM \cite{sam1} decoder. 
SasP employs a parameter-free similarity-guided approach in which it calculates the semantic similarity between the \texttt{<SEG}> token embedding and the image token embedding to identify relevant regions. 
It then converts similarity scores into differentiable point prompts through a Gaussian-like weighted average interpolation, allowing gradients to flow back through the similarity maps during training. 
This design enables simultaneously ``reason to attend'' during forward passes and ``attend to reason'' during backpropagation. 
Unlike previous approaches \cite{lisa,lisa++,lasagna,gsva} that rely solely on the \texttt{<SEG}> token for mask generation, READ actively guides the model's attention using similarity-derived points.

\paragraph{GLaMM}
Grounding LMM (GLaMM) \cite{glamm} aims to address the limitations that previous image RS methods lack abilities of multi-round conversation and processing image-wise prompting. 
GLaMM is an end-to-end model, consisting of five components, namely a global image encoder, a region encoder, an LLM, a grounding image encoder (based on SAM \cite{sam1}), and a pixel decoder to generate natural language responses together with the corresponding segmentation masks in response to implicit text queries and optional visual prompts (\textit{e}.\textit{g}., bounding box). 
The global and region encoders in GLaMM enable scene-level and region-specific understanding, respectively, while the grounding image encoder and pixel decoder facilitate fine-grained pixel-level grounding without the need for external segmentation models. 
Unlike LISA \cite{lisa} and its variants, which are limited to single-target scenarios, GLaMM can handle multiple targets simultaneously within a single conversation.

\paragraph{SESAME}
SESAME (SEe, SAy, segMEnt) \cite{sesame} aims to handle false-premise queries where referenced objects do not exist in the image. 
Unlike previous approaches that tend to hallucinate masks for non-existent objects, SESAME enables three essential capabilities.
First, it can ``see'' by detecting whether queried objects exist in the image. 
Second, it can ``say'' by providing informative feedback when objects are absent, providing clarifications, suggesting alternative expressions, or correcting semantic errors in queries. 
Third, it can ``segment'' by generating masks for objects that do exist. 
SESAME is trained on a unified dataset that combines positive referring expressions, false-premise queries with corrections, and visual question-answer data to prevent catastrophic forgetting of conversational abilities during segmentation training. 
SESAME differs from other image RS methods, such as LISA \cite{lisa} and GSVA \cite{gsva}, which primarily focus on segmentation without addressing false premises or only focusing on object absence, which lack the ability to engage in natural dialogue.

\begin{figure}[htbp!]
\centering
\includegraphics[width=1.0\linewidth]{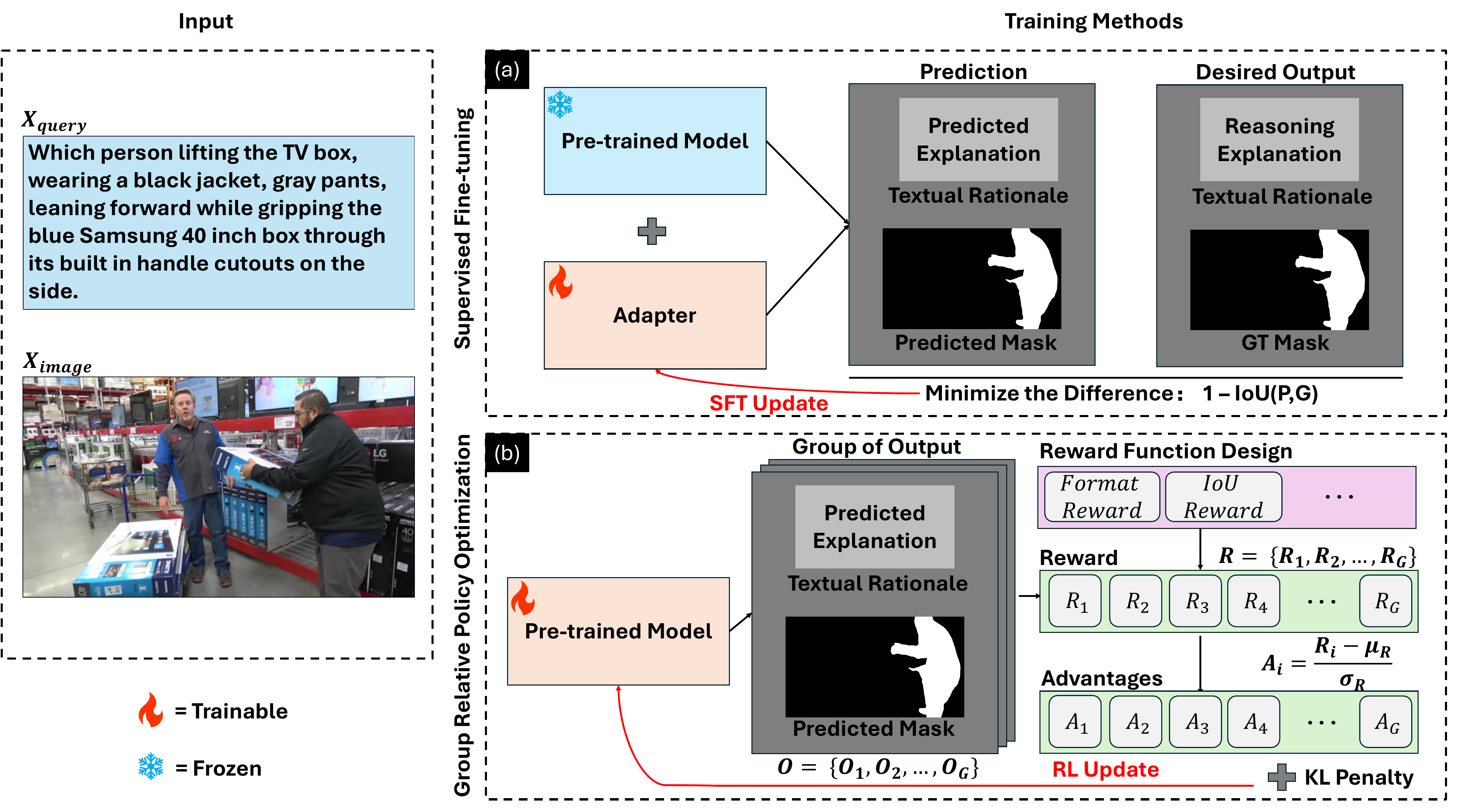}
\caption{
Comparison of training approaches for RS models: (a) Supervised fine-tuning as in LISA and variants \cite{lisa}, and (b) Reinforcement learning optimizes the model using a reward function that combines format rewards and accuracy rewards, as implemented in Seg-Zero \cite{segzero}. 
}\label{fig:train}
\end{figure}

\paragraph{Seg-Zero}
Seg-Zero \cite{segzero} tackles limitations of other image RS methods by focusing on outside-domain generalization and showing explicit reasoning processes for improved performance, as shown in Fig.~\ref{fig:train}. 
Specifically, Seg-Zero introduces a decoupled architecture consisting of a reasoning model (\textit{i}.\textit{e}., MLLM) and a segmentation model (\textit{i}.\textit{e}., SAM) trained exclusively through reinforcement learning with Generalized Reward Policy Optimization (GRPO). 
The reasoning model interprets implicit text queries, generates explicit chain-of-thought reasoning, and finally generates positional prompts (bounding boxes and points) which are subsequently used by the segmentation model to generate precise pixel-level masks. 
The reward integrates both format rewards (enforcing structured reasoning output) and accuracy rewards (based on IoU of segmentation and L1 distance metrics of the prompt) to guide GRPO optimization of reasoning model.
Trained without explicit reasoning data, Seg-Zero exhibits test-time reasoning capabilities and improved generalization across domains.

\paragraph{MIRAS}
MIRAS \cite{miras} enables fine-grained image RS with a pixel-level focus through multi-turn conversational interactions and progressive reasoning that tracks evolving user intent. 
The MIRAS framework consists of three components: a visual encoder, a MLLM, and a mask decoder, connected by a special \texttt{<SEG>} token for end-to-end processing. 
MIRAS incorporates dual visual encoders that extract enriched visual features at different resolutions, and a semantic region alignment strategy to refine the model's focus by incorporating target semantic information. 
This architecture design effectively combines conversational reasoning with pixel-level precision, allowing the model to generate both pixel-grounded explanations and accurate segmentation masks. 
Trained through a two-stage process of mask-text alignment pre-training followed by instruction fine-tuning, MIRAS achieves robust performance in both segmentation accuracy and reasoning quality.

\paragraph{POPEN}
POPEN \cite{popen} addresses two limitations in existing LVLM-based image RS approaches: imprecise segmentation results and hallucinations in text responses.
Unlike most previous methods that rely solely on supervised fine-tuning, POPEN incorporates preference-based optimization to align the LVLM with human preferences through reinforcement learning. 
Moreover, it introduces a curriculum learning mechanism by focusing first on object localization, then on boundary refinement. 
To further enhance performance, POPEN also implements a preference-based ensemble method that integrates multiple outputs from the LVLM.

\subsection{Evaluation Metrics}

We present an overview comparison of all the image RS metrics in Table~\ref{table:rs_metrics}.

\begin{table*}[htbp]
\caption{
Comparison of evaluation metrics for image RS across multiple dimensions, including their primary focus, mathematical formulation, 
original purpose/domain, advantages, limitations, implementation complexity, and usage frequency in RS literature.
}
\label{table:rs_metrics}
\centering
\resizebox{\linewidth}{!}{
\begin{tabular}{l|c|c|c|c|c|c|c}
\toprule
Metric & Primary Focus & Formula & Original Purpose/Domain & Advantages & Limitations & Implementation Complexity & Usage Frequency \\
\hline
\rowcolor{gray!5}
cIoU & Mask Quality & $\frac{\sum_{i=1}^{N} |P_i \cap G_i|}{\sum_{i=1}^{N} |P_i \cup G_i|}$ & 
Object detection, segmentation & 
Robust to small objects; aggregates before dividing & 
Favors large objects; biased toward high-density regions & 
Low & 
Very Common \\
ncIoU & Mask Quality & $\frac{\sum_{i=1}^{N} |R(P_i) \cap R(G_i)|}{\sum_{i=1}^{N} |R(P_i) \cup R(G_i)|}$ & 
RS with varying image sizes & 
Size-invariant; equalizes contribution across images & 
Additional preprocessing step; may lose resolution details & 
Medium & 
Common \\
\rowcolor{gray!5}
gIoU & Mask Quality & $\frac{1}{N}\sum_{i=1}^{N} \frac{|P_i \cap G_i|}{|P_i \cup G_i|}$ & 
Object detection, bounding box regression & 
Instance-level assessment; differentiates non-overlapping cases & 
Sensitive to outliers; can be less stable than cIoU & 
Low & 
Common \\
mIoU & Mask Quality & $\frac{1}{K}\sum_{k=1}^{K} \frac{|P_k \cap G_k|}{|P_k \cup G_k|}$ & 
Semantic segmentation & 
Class-aware; handles imbalanced datasets; equalizes class contribution & 
Requires class label annotations; less sensitive to rare classes & 
Medium & 
Very Common \\
\rowcolor{gray!5}
sIoU & Text Reasoning & $\frac{|\hat{c} \cap y|}{|\hat{c} \cup y|}$ & 
Text-based evaluation in RS & 
Word-level precision evaluation; simple to implement & 
Only considers exact matches; ignores semantic similarities & 
Low & 
Occasional \\
MUSE & Both & Multi-step process with Hungarian matching & 
Multi-target RS & 
Jointly evaluates mask and text quality; handles multiple targets & 
Complex implementation; requires GPT-3.5 for quality assessment & 
High & 
Emerging \\
\rowcolor{gray!5}
SS & Text Reasoning & $\langle g(\hat{c}), g(y) \rangle$ & 
Semantic comparison in NLP & 
Captures semantic relationships beyond exact matches & 
Embedding quality dependent; requires pretrained embedding models & 
Medium & 
Occasional \\
CHAIR & Text Reasoning & $\text{CS} = \frac{|\text{responses with hallucinated objects}|}{|\text{all responses}|}, \text{CI} = \frac{|\text{hallucinated objects}|}{|\text{all mentioned objects}|}$ & 
Image captioning & 
Quantifies hallucination at instance and sentence levels & 
Limited to predefined object categories; requires ground truth objects & 
High & 
Emerging \\
\rowcolor{gray!5}
GPT-score & Text Reasoning & LLM-based evaluation (1-10 scale) & 
Comprehensive text evaluation & 
Holistic assessment; considers reasoning quality and factual accuracy & 
Subjective; dependent on LLM quality; reproducibility challenges & 
High & 
Emerging \\
\bottomrule
\end{tabular}
}
\end{table*}
\paragraph{cIoU}
The cumulative Intersection over Union (cIoU) aims to evaluate RS performance at the dataset level \cite{lisa}. 
For a given prediction mask $P$ and the corresponding ground truth mask $G$, the Intersection over Union (IoU) is calculated as the ratio of their intersection to their union, \textit{i}.\textit{e}. 
\begin{equation}
\text{IoU} = \frac{|P \cap G|}{|P \cup G|}.
\end{equation}
The cIoU extends IoU by aggregating intersection and union values across all samples before computing their ratio:
\begin{equation}
\text{cIoU} = \frac{\sum_{i=1}^{N} |P_i \cap G_i|}{\sum_{i=1}^{N} |P_i \cup G_i|},
\end{equation}
where $N$ represents the total number of samples in the dataset, and $P_i$ and $G_i$ denote the prediction and ground truth masks for the $i$-th sample, respectively. 
This cumulative computation can prevent individual samples with small regions from disproportionately affecting the overall performance metric.

\paragraph{ncIoU} 
The cIoU ignores the contribution of the image size; therefore, a normalized cumulative IoU (ncIoU) is proposed \cite{llmseg}.
ncIoU first standardizes each prediction and ground truth mask to a uniform size before computing the intersection and union statistics. 
Formally, given a set of $N$ image pairs where $P_i$ and $G_i$ represent the prediction and ground truth masks for the $i$-th sample respectively, ncIoU is computed as:
\begin{equation}
\text{ncIoU} = \frac{\sum_{i=1}^{N} |R(P_i) \cap R(G_i)|}{\sum_{i=1}^{N} |R(P_i) \cup R(G_i)|},
\end{equation}
where $R(\cdot)$ denotes the resizing operation that normalizes each prediction and ground truth mask to a standard resolution. 
This normalization ensures that each image pair contributes equally to the final metric regardless of its original dimensions.

\paragraph{gIoU}
The generalized Intersection over Union (gIoU) offers a complementary perspective to cIoU by evaluating RS performance at the instance level \cite{lisa,giou}. 
Unlike cIoU, which provides a dataset-level assessment and favors larger objects \cite{wu2020phrasecut,yang2022lavt}, gIoU computes the mean IoU across all samples, treating each instance equally regardless of its size, formulated as
\begin{equation}
\text{gIoU} = \frac{1}{N}\sum_{i=1}^{N} \frac{|P_i \cap G_i|}{|P_i \cup G_i|}.
\end{equation}
This metric can, therefore, evaluate the RS model's consistency across different instances and scales, as it gives equal weight to each prediction regardless of the object's size. 
In cases where the ground truth mask is empty (\textit{i}.\textit{e}., indicating that there is no target object), and the model correctly predicts no object ($|P_i \cup G_i| = 0$), the instance is assigned a perfect score of $\frac{|P_i \cap G_i|}{|P_i \cup G_i|}=1.0$ to reward accurate negative predictions.

\paragraph{mIoU}
The mean Intersection-over-Union (mIoU) is widely used for semantic segmentation evaluation \cite{segllm,lasagna}, which can be used in the case of RS with multiple semantic categories. 
Unlike gIoU, which computes the mean IoU across all samples, mIoU calculates the IoU for each semantic class independently and then averages across all classes. 
Specifically, for each semantic class $k$, the mIoU first computes the IoU between the predicted segmentation mask $P$ and the ground truth mask $G$, namely $\text{IoU}_k = \frac{|P_k \cap G_k|}{|P_k \cup G_k|}$, where $P_k$ and $G_k$ denote the binary masks for the semantic class $k$ in the prediction and ground truth, respectively. 
The final mIoU is then obtained by averaging these semantic class-specific IoUs, formulated as
\begin{equation}
\text{mIoU} = \frac{1}{K}\sum_{k=1}^{K} \text{IoU}_k,
\end{equation}
where $K$ is the total number of semantic classes. 
This class averaging makes mIoU suitable for evaluating performance across multiple semantic categories, as it gives equal weight to each class regardless of their frequency or pixel area in the dataset. 
This is in contrast to gIoU, which focuses on instance-level evaluation by computing the mean IoU across all samples without consideration of class boundaries.

\paragraph{sIoU}
Semantic IoU (sIoU) \cite{siou,prima} measures the overlap between the words in predicted and ground-truth class names, which also serves as an evaluation metric for the reasoning capabilities of the generated text.
For a prediction $\hat{c}$ and ground truth $y$, treating each as a set of words, the sIoU is calculated as $\text{sIoU} = |{\hat{c} \cap y}|/|{\hat{c} \cup y}|$, where $|\cdot|$ denotes the cardinality of the set. 
sIoU penalizes predictions that include irrelevant terms or miss terms from the ground truth label. 
For example, if the ground truth is ``\textit{red car}'' and the prediction is ``\textit{blue car}'', the sIoU would be $0.5$ since only one word overlaps. 
The sIoU provides a complementary perspective to semantic similarity by focusing on exact word-level precision rather than general semantic relatedness. 
When the ground truth mask is empty (\textit{i}.\textit{e}., indicating that there is no target object) and the model correctly predicts no object ($|\hat{c} \cup y| = 0$), the instance is assigned a perfect score of $1.0$ to reward accurate negative predictions.

\paragraph{MUSE}
The Multi-target Use Segmentation (MUSE) metric \cite{pixellm} is proposed specifically for RS models that can generate both text descriptions and the corresponding masks for multiple targets. 
Unlike IoU and variants that focus solely on mask quality, MUSE assesses both mask accuracy and the alignment between text descriptions and visual segments. 
The evaluation metric process involves four steps.
Firstly, predicted masks are matched with ground-truth masks using bipartite matching via the Hungarian algorithm, with unassigned elements receiving empty masks. 
Secondly, the generated text is modified by replacing mask descriptions with the corresponding ground-truth descriptions to standardize the textual content. 
Thirdly, a GPT-3.5 assigns a quality score ($s_i$) to each prediction based on how well it aligns with the description. 
Finally, the IoU for each prediction is computed and weighted by the quality score, where:
\begin{equation}
\text{Intersection}_i = \begin{cases}
\text{Intersection}_i & \text{if } s_i > 0.5 \\
0 & \text{if } s_i \leq 0.5
\end{cases}.
\end{equation}
The final image-level IoU is calculated as the sum of all prediction IoUs divided by the total number of predictions, namely
\begin{equation}
\text{IoU}_i = \frac{\text{Intersection}_i}{\text{Union}_i}.
\end{equation}
MUSE's advantage over metrics like cIoU, ncIoU, and gIoU is its ability to evaluate multi-target RS while considering both mask quality and semantic alignment with text descriptions.

\paragraph{SS}
Semantic Similarity (SS) \cite{siou,prima} evaluates how well the predicted class name aligns with the ground-truth label in a semantic embedding space, which serves as an evaluation metric for the reasoning capabilities of the generated text responses in RS. 
Given an input image $x$ with ground-truth label $y$ and predicted class $\hat{c} = f(x)$, the semantic similarity is computed as 
\begin{equation}
\text{SS} = \langle g(\hat{c}), g(y) \rangle, 
\end{equation}
where $g: \mathcal{T} \rightarrow \mathcal{Y}$ maps text to an embedding space $\mathcal{Y}$ using Sentence-BERT \cite{reimers2019sentence}, and $\langle\cdot\rangle$ is the inner product. 
This metric captures semantic relationships between text predictions and ground truth that might not be apparent from exact string matching. 
The embedding space $\mathcal{Y}$ allows for the comparison of free-form text descriptions, making it suitable for RS where predictions may use varied but semantically related terminology.

\paragraph{CHAIR}
The CHAIR (Caption Hallucination Assessment with Image Relevance) metric \cite{popen,chair} evaluates hallucination in RS by measuring the proportion of objects mentioned in the generated text response that are absent from the ground truth.
CHAIR includes two sub-metrics: CS (Caption-level Score) and CI (Instance-level Score), formulated as:
\begin{equation}
\text{CS} = \frac{|{\text{responses w/ hallucinated objects}}|}{|{\text{all responses}}|}, \quad \text{CI} = \frac{|{\text{hallucinated objects}}|}{|{\text{all mentioned objects}}|}.
\end{equation}
CS measures the proportion of responses that contain at least one hallucinated object, while CI calculates the ratio of hallucinated objects to all objects mentioned in the responses.
Unlike previous IoU-based metrics that focus solely on segmentation quality, CHAIR evaluates the semantic accuracy of textual descriptions generated in RS tasks, providing a complementary perspective for applications requiring accurate segmentation and hallucination-free text generation.

\paragraph{GPT-score}
The GPT score \cite{popen,huang2024opera} assesses the quality of the text response in RS using LLMs such as ChatGPT to evaluate responses.
This metric prompts ChatGPT to assess the correctness of an LVLM's response given the input image-instruction pair, with particular attention to hallucinations.
The evaluation considers two primary criteria: (1) the accuracy of the response in relation to image content and reasoning logic, rewarding responses with fewer hallucinations, and (2) the completeness and detail level of the response.
The resulting score ranges from 1 to 10, with higher scores indicating better performance.

\begin{table*}[t!]
\caption{
Comparison of RS datasets. 
We characterize datasets across multiple dimensions, including the annotation approach, query complexity, and data scale. 
The modality (``Mo.'') column indicates the vision data type, while annotation characteristics encompass the method of annotation (``Ann. Method''), use of automated tools (``Auto.''), support for multi-round conversations (``Conv.''), and type of segmentation mask. 
Query properties describe the linguistic characteristics of text inputs, including length, reasoning complexity (``Low'': simple referring expressions, ``High:'' complex reasoning chains, ``Very High'': multi-step or comparative reasoning), and language style (``Natural'': human-written descriptions, ``Template'': structured patterns, ``Free-form'': LLM-generated diverse expressions). 
For M\textsuperscript{4}Seg, asterisks (*) indicate the number of question-answer pairs rather than images.
%
}\label{table:rs_dataset}
\centering
\resizebox{\linewidth}{!}{
\begin{tabular}{l|c|c|cccc|ccc|cccc|c}
\toprule
\multirow{2}{*}{Dataset} & \multirow{2}{*}{Mo.} & \multirow{2}{*}{\makecell[c]{Image\\Source}} & \multicolumn{4}{c|}{Annotation Characteristics} & \multicolumn{3}{c|}{Query Properties} & \multicolumn{4}{c|}{Scale (\# Data)} & \multirow{2}{*}{\makecell[c]{Publicly\\Available}} \\
\cline{4-14}
& & & Ann. Method & Auto. & Conv. & Mask Type & Length & Reasoning & Language & Train & Val & Test & Total & \\
\hline
\rowcolor{gray!5}
RefCOCO (google) \cite{refcoco} & \cellimage{figs/icon/img.pdf} & MS COCO \cite{mscoco} & Manual & \xmark & \xmark & Semantic & Short & Low & Natural & 19,213 & 4,559 & 4,527 & 28,299 & \cmark \\
RefCOCO (unc) \cite{refcocounc} & \cellimage{figs/icon/img.pdf} & MS COCO \cite{mscoco} & Manual & \xmark & \xmark & Semantic & Short & Low & Natural & 16,994 & 1,500 & 1,500 & 19,994 & \cmark \\
\rowcolor{gray!5}
RefCOCO+ \cite{refcoco} & \cellimage{figs/icon/img.pdf} & MS COCO \cite{mscoco} & Manual & \xmark & \xmark & Semantic & Short & Low & Natural & 16,992 & 1,500 & 1,500 & 19,992 & \cmark \\
RefCOCOg (google) \cite{refcocog} & \cellimage{figs/icon/img.pdf} & MS COCO \cite{mscoco} & Manual & \xmark & \xmark & Semantic & Medium & Low & Natural & 24,698 & 4,650 & 0 & 29,348 & \cmark \\
\rowcolor{gray!5}
RefCOCOg (umd) \cite{refcocog} & \cellimage{figs/icon/img.pdf} & MS COCO \cite{mscoco} & Manual & \xmark & \xmark & Semantic & Medium & Low & Natural & 21,899 & 1,300 & 2,600 & 25,799 & \cmark \\
ReasonSeg \cite{lisa} & \cellimage{figs/icon/img.pdf} & OpenImages \cite{openimages}, ScanNetv2 \cite{scannet} & Manual & \xmark & \xmark & Semantic & Mixed & High & Template & 239 & 200 & 799 & 1,238 & \cmark \\
\rowcolor{gray!5}
ReasonSeg-Inst \cite{lisa++} & \cellimage{figs/icon/img.pdf} & MS COCO \cite{coco}, ADE20K \cite{ade20k} & GPT-4V & \cmark & \xmark & Instance & Long & High & Free-form & 62,000 & 0 & 1,800 & 63,800 & \cmark \\
LLM-Seg40K \cite{llmseg} & \cellimage{figs/icon/img.pdf} & LVIS \cite{lvis}, EgoObjects \cite{egoobjects} & LLaVA+GPT-4 & \cmark & \xmark & Semantic & Long & High & Free-form & 11,000 & 1,000 & 2,000 & 14,000 & \cmark \\
\rowcolor{gray!5}
%
MRSeg \cite{segllm} & \cellimage{figs/icon/img.pdf} & Multiple Sources & GPT-4 & \cmark & \cmark & Semantic & Long & High & Free-form & 413,877 & 41,856 & 0 & 455,733 & \cmark \\
MRSeg (hard) \cite{segllm} & \cellimage{figs/icon/img.pdf} & MRSeg \cite{segllm} & GPT-4 & \cmark & \cmark & Semantic & Long & Very High & Free-form & 22,470 & 1,988 & 0 & 24,458 & \cmark \\
\rowcolor{gray!5}
M\textsuperscript{4}Seg \cite{prima} & \cellimage{figs/icon/img.pdf} & Multiple Sources & GPT-4 & \cmark & \xmark & Semantic & Long & Very High & Free-form & 224,393* & 0 & 5,000* & 229,393* & \xmark \\
MUSE \cite{pixellm} & \cellimage{figs/icon/img.pdf} & LVIS \cite{lvis} & GPT-4V & \cmark & \xmark & Instance & Long & High & Free-form & 239,000 & 2,800 & 4,300 & 246,100 & \cmark \\
\rowcolor{gray!5}
GranD \cite{glamm} & \cellimage{figs/icon/img.pdf} & SAM images \cite{sam1} & Automated pipeline & \cmark & \cmark & Instance & Long & Very High & Free-form & 11M & - & - & 11M & \cmark \\
GranD\textsubscript{f} \cite{glamm} & \cellimage{figs/icon/img.pdf} & Multiple Sources & GPT-4 + Manual & \cmark & \cmark & Instance & Long & Very High & Free-form & - & 2,500 & 5,000 & 214,000 & \cmark \\
\rowcolor{gray!5}
FP-RefCOCO \cite{sesame} & \cellimage{figs/icon/img.pdf} & RefCOCO \cite{refcoco} & GPT-3.5-turbo & \cmark & \xmark & Semantic & Short & High & Free-form & 16,992 & 1,500 & 1,500 & 19,992 & \cmark \\
FP-RefCOCO+ \cite{sesame} & \cellimage{figs/icon/img.pdf} & RefCOCO+ \cite{refcoco} & GPT-3.5-turbo & \cmark & \xmark & Semantic & Short & High & Free-form & 16,994 & 1,500 & 1,500 & 19,994 & \cmark \\
\rowcolor{gray!5}
FP-RefCOCOg \cite{sesame} & \cellimage{figs/icon/img.pdf} & RefCOCOg \cite{refcocog} & GPT-3.5-turbo & \cmark & \xmark & Semantic & Medium & High & Free-form & 21,899 & 1,300 & 2,600 & 25,799 & \cmark \\
PRIST \cite{miras} & \cellimage{figs/icon/img.pdf} & TextCaps \cite{sidorov2020textcaps} & GPT-4o + Manual & \cmark & \cmark & Pixel-level & Long & Very High & Free-form & 7,312 & 500 & 508 & 8,320 & \cmark \\
\rowcolor{gray!5}
MMR \cite{jang2025mmr} & \cellimage{figs/icon/img.pdf} & PACO-LVIS \cite{ramanathan2023paco} & GPT-4V + Manual & \cmark & \cmark & Multi-granularity & Long & Very High & Free-form & 154,127 & 8,194 & 32,077 & 194,398 & \cmark \\
\hline
Refer-Youtube-VOS \cite{seo2020urvos} & \cellimage{figs/icon/video.pdf} & Youtube-VOS \cite{xu2018youtube} & Manual & \xmark & \xmark & Semantic & Mixed & Low & Natural & 23,810 & 4,089 & 0 & 27,899 & \cmark \\
\rowcolor{gray!5}
Refer-DAVIS\textsubscript{17} \cite{khoreva2019video} & \cellimage{figs/icon/video.pdf} & DAVIS17 \cite{pont20172017} & Manual & \xmark & \xmark & Semantic & Mixed & Low & Natural & - & - & - & $\sim$1.2k & \cmark \\
InsTrack \cite{trackgpt} & \cellimage{figs/icon/video.pdf} & Refer-DAVIS17 \cite{khoreva2019video}, VIPSeg \cite{miao2022large} & Manual+ChatGPT & \xmark & \xmark & Semantic & Mixed & High & Natural & 858 & 240 & 0 & 1098 & \cmark \\
MeViS \cite{ding2023mevis} & \cellimage{figs/icon/video.pdf} & Multiple Sources \cite{ding2023mose,qi2022occluded,wang2021unidentified,voigtlaender2021reducing} & Manual & \xmark & \xmark & Semantic & Medium & High & Natural & 1,712 & 140 & 154 & 2,006 & \cmark \\
\rowcolor{gray!5}
ReVOS \cite{visa} & \cellimage{figs/icon/video.pdf} & Multiple Sources \cite{wang2024ov,wang2023towards,ding2023mose,qi2022occluded,dave2020tao,wang2021unidentified} & Manual & \xmark & \xmark & Semantic & Long & High & Natural & 626 & 416 & 0 & 1,042 & \cmark \\
VideoReasonSeg \cite{villa} & \cellimage{figs/icon/video.pdf} & YouTube-VIS \cite{yang2019video}, OVIS \cite{qi2022occluded}, LV-VIS \cite{wang2023towards} & GPT-4V & \cmark & \xmark & Semantic & Long & Very High & Free-form & 1,000 & 400 & 534 & 1,934 & \cmark \\
\rowcolor{gray!5}
ReasonVOS \cite{videolisa} & \cellimage{figs/icon/video.pdf} & Multiple Sources \cite{ding2023mose,ding2023mevis,miao2022large,athar2023burst} & LLM+Manual & \cmark & \xmark & Semantic & Mixed & Very High & Free-form & 0 & 0 & 458 & 458 & \cmark \\
GroundMORE \cite{mora} & \cellimage{figs/icon/video.pdf} & YouTube & GPT-4o + Manual & \cmark & \xmark & Spatiotemporal & Long & Very High & Free-form & 1,333 & 0 & 382 & 1,715 & \cmark \\
\rowcolor{gray!5}
ViCaS \cite{vicas} & \cellimage{figs/icon/video.pdf} & Oops \cite{epstein2020oops} & SAM2 \cite{sam2} + Manual & \cmark & \xmark & Instance & Long & High & Natural & 5,131 & 1,100 & 1,100 & 7,331 & \cmark \\
JiT Benchmark \cite{jitdt} & \cellimage{figs/icon/video.pdf} & DAVIS \cite{pont20172017}, SA-V \cite{sam2} & Manual & \xmark & \xmark & Semantic & Long & Very High & Free-form & 0 & 0 & 895 & 895 & \cmark \\
\bottomrule
\end{tabular}
}
\end{table*}

\subsection{Datasets and Benchmarks}

\begin{figure}[htbp!]
    \centering
    \includegraphics[width=1.0\linewidth]{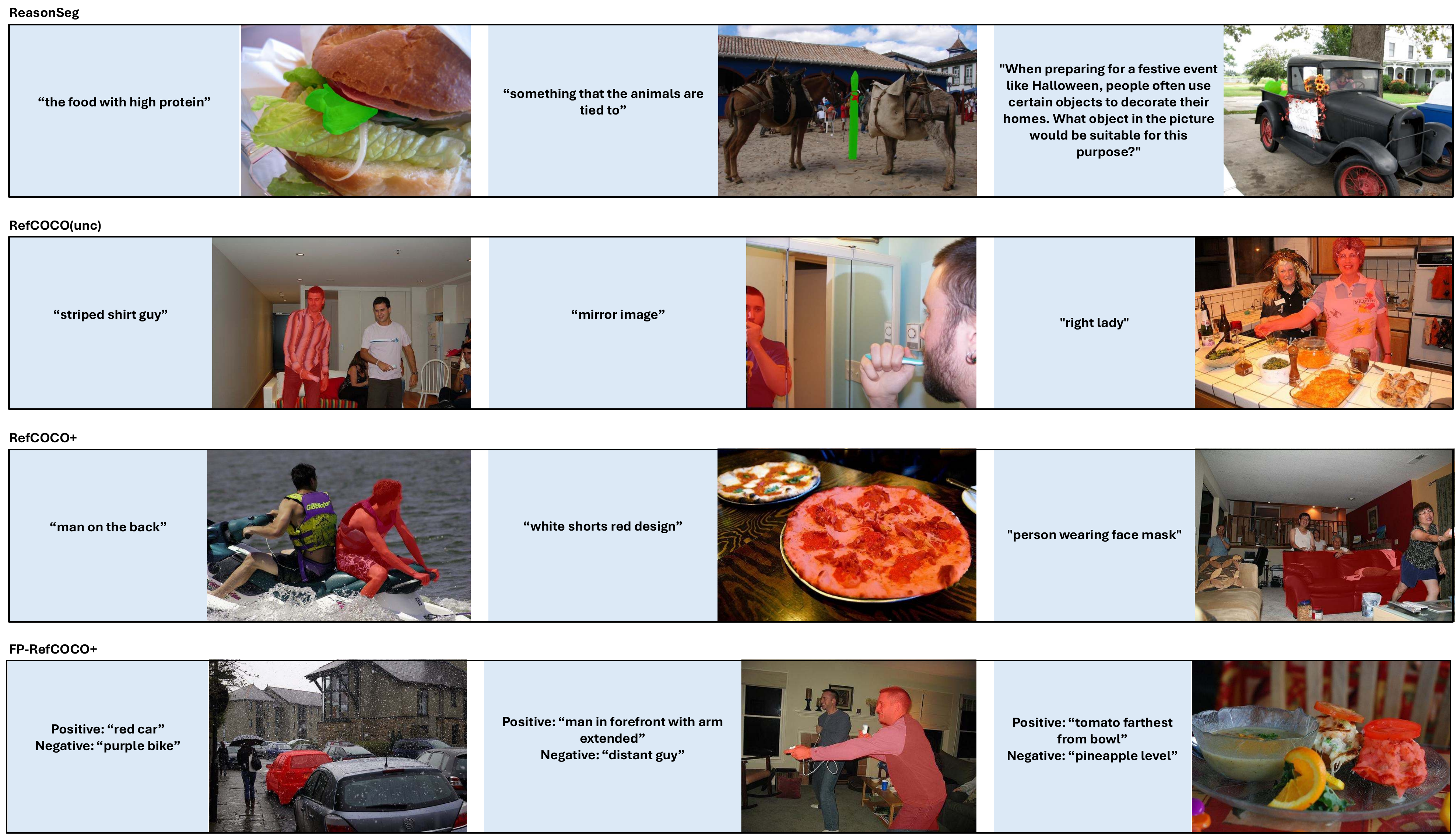}
    \caption{
    Illustrative example of image RS data.
    }\label{fig:benchmark1}
\end{figure}

Table \ref{table:rs_dataset} presents a comparison of RS datasets in multiple dimensions, including annotation methodologies, query properties, and data scale. 
The evolution of RS datasets reflects a clear progression from simple referring expressions to increasingly complex reasoning tasks. 
Early datasets like RefCOCO and RefCOCO+ \cite{refcoco} established foundational benchmarks with manually annotated natural language descriptions. 
More recent datasets, such as ReasonSeg \cite{lisa} and M\textsuperscript{4}Seg \cite{prima}, have introduced more complex reasoning challenges and leveraged LLMs/MLLMs for annotation, while datasets like MRSeg \cite{segllm} have explored multi-round conversational interactions. 
Fig.~\ref{fig:benchmark1} provides some examples of image RS data.

\paragraph{RefCOCO and RefCOCO+}
RefCOCO and RefCOCO+ \cite{refcoco,refcocounc} were initially constructed for referring expression generation and later used in referring segmentation.
In the field of RS, they are also widely adopted as a benchmark, although their text queries are relatively simple compared to other RS-specific benchmarks.
RefCOCO contains 142,209 text expressions that describe 50,000 objects in 19,994 images of the MS COCO dataset, collected through an interactive two-player game (\textit{i}.\textit{e}., ReferitGame) \cite{refcoco} where participants alternate between describing and identifying objects based on text descriptions.
It was first constructed on the 20,000 images from the ImageCLEF dataset \cite{grubinger2006iapr}, and was extended with images from MS COCO \cite{mscoco}.
RefCOCO has two widely used data splits: 
1) UNC split \cite{refcocounc} partitions images into train/val/test sets, 
2) Google split separates object instances within images across splits, allowing image reuse with different referred objects.
RefCOCO+ extends RefCOCO with stricter annotation guidelines, prohibiting spatial descriptors to focus exclusively on appearance-based references \cite{refcoco,refcocounc}.
It has 141,564 text expressions for 49,856 objects in 19,992 images.
It shares the UNC split with RefCOCO, but removes location-dependent language through explicit annotation constraints.

\paragraph{RefCOCOg}
The RefCOCOg \cite{refcocog} dataset extends the RefCOCO \cite{refcoco} with a focus on richer linguistic descriptions. 
Collected through a non-interactive annotation process rather than the game-based approach of its predecessors \cite{refcoco}, it contains 95,010 referring expressions across 25,799 MS COCO \cite{mscoco} images.
RefCOCOg has two splits: 1) Google split with 24,698 training and 4,650 validation images but no test set, and 2) UMD split (21,899 train/1,300 val/2,600 test images) that strictly separate images across partitions to prevent data leakage.

\paragraph{ReasonSeg}
ReasonSeg \cite{lisa} is the first benchmark specifically designed for RS.
It consists of 1,218 image-instruction-mask samples, where the images are sourced from OpenImages \cite{openimages} and ScanNetv2 \cite{scannet}. 
Each sample contains an input image, a manually annotated implicit text query that requires reasoning, and a corresponding manually annotated ground-truth segmentation mask. 
The implicit text queries in ReasonSeg can be categorized into two main types: 
(1) short phrases and (2) long sentences. 
Short phrase queries typically focus on object functionality or implicit properties (\textit{e}.\textit{g}., ``\textit{the camera lens that is more suitable for photographing nearby objects}''), while long sentence queries often involve more complex reasoning chains about safety, utility, or spatial relationships (\textit{e}.\textit{g}., ``\textit{After cooking, consuming food, and preparing for food, where can we throw away the rest of the food and scraps?}'').
To facilitate both the evaluation and fine-tuning of MLLM, ReasonSeg is divided into three splits: (1) a training set of 239 samples, (2) a validation set of 200 samples, and (3) a test set of 779 samples.

\paragraph{ReasonSeg-Inst}
The ReasonSeg-Inst dataset \cite{lisa++} aims to extend the ReasonSeg dataset \cite{lisa} with instance-level masks. 
To be more specific, it focuses on differentiating between multiple instances of the same semantic category in RS. 
The images and instance segmentation masks in ReasonSeg-Inst were curated directly from COCO \cite{coco} and ADE20K \cite{ade20k}, with samples filtered to ensure high-quality data and avoid impeding the ability of GPT-4V \cite{gpt4v} in synthesizing text pairs.
Specifically, images from COCO and ADE20K smaller than $512\times512$ pixels or containing objects with areas under 400 square pixels were excluded. 
Instead of manual annotation as conducted in ReasonSeg \cite{lisa}, ReasonSeg-Inst uses GPT-4V \cite{gpt4v} to generate query-answer pairs about instances given the image. 
It comprises 62,000 samples for training and 1,800 samples for testing. 
Each sample consists of an input image, an instance-aware implicit text query, a corresponding text answer, and ground-truth segmentation masks.

\paragraph{LLM-Seg40K}
The LLM-Seg40K construction \cite{llmseg} leverages images and corresponding semantic segmentation masks or bounding box annotations from LVIS \cite{lvis} and EgoObjects \cite{egoobjects}. 
To ensure the diversity of data in LLM-Seg40K, while maintaining the complexity of implicit text queries, LLM-Seg40K selected 6,000 photographic images with 2-5 different semantic categories (\textit{i}.\textit{e}., simple images) and 4,000 images with more than 5 semantic categories (\textit{i}.\textit{e}., complex images) from LVIS, complemented by 3,000 egocentric images with more than 2 semantic categories from EgoObject. 
For LVIS samples, the segmentation mask with respect to the given semantic category is already available.
For EgoObjects samples, the SAM \cite{sam1} was used to convert the bounding box into precise binary segmentation masks.
The annotation process for implicit text queries was automated by MLLMs.
Initially, LLaVA-v1.5-13B \cite{llava} generated detailed image descriptions. 
These descriptions, along with mask semantic categories, were then fed into GPT-4 \cite{gpt4}, prompted to generate the final implicit text queries. 
Each image is decorated with multiple implicit text quires, with an average of 3.95 questions per image and an average question length of 15.2 words. 
LLM-Seg40K is divided into training sets (11,000 images), validation sets (1,000 images), and testing sets (2,000 images) on the image level. 

\paragraph{MRSeg}
The multi-round RS segmentation (MRSeg) dataset \cite{segllm} aggregates and augments data from RefCOCO+ \cite{refcoco}, RefCOCOg \cite{refcocog}, Visual Genome \cite{krishna2017visual}, PACO-LVIS \cite{ramanathan2023paco}, LVIS \cite{lvis}, Pascal Panoptic Part \cite{de2021part}, ADE20K \cite{ade20k}, COCO-Stuff \cite{caesar2018coco}, and MS COCO \cite{mscoco}, where they use the bounding box or segmentation annotations from these original datasets to generate multi-round natural language conversations.
For queries involving hierarchical relationships, where conversations begin with queries about whole objects before progressing to their constituent parts, it leverages images and corresponding masks from PACO-LVIS \cite{ramanathan2023paco} and Pascal Panoptic Part \cite{de2021part}. 
Each conversation includes between one to four instances, and each instance has one to four associated parts. 
For queries involving positional relationships, based on RefCOCO+ \cite{refcoco}, RefCOCOg \cite{refcocog}, and LVIS \cite{lvis}, MRSeg incorporates spatial relationships between objects into the implicit text queries, sampling between 2 to 18 annotations per image. 
The MRSeg dataset also includes samples to evaluate interactional relationships based on Visual Genome \cite{krishna2017visual}, with up to four relationship annotations per image structured as two-round conversations.
The training set comprises a substantial collection of conversations across different source datasets: 55,188 conversations from RefCOCO+ \cite{refcoco} and RefCOCOg \cite{refcocog}, 367,674 from Visual Genome \cite{krishna2017visual}, 40,827 from PACO-LVIS \cite{ramanathan2023paco}, 71,388 from LVIS \cite{lvis}, and 4,577 from Pascal Panoptic Part \cite{de2021part}. 
Additionally, it includes images with single-round conversations, namely 59,784 from ADE20K \cite{ade20k}, 340,127 from COCO-Stuff \cite{coco}, and 49,036 attribute-based conversations from MS COCO \cite{mscoco}. 
The validation set maintains a similar distribution but with a smaller scale.
To ensure the natural language quality of text queries, GPT-4 \cite{gpt4} is used to generate various question templates for each type of relationship and to refine the grammatical structure of conversations. 
MRSeg also includes a challenging subset, MRSeg (hard), specifically designed to evaluate models' ability to understand and utilize information from previous conversation rounds. 
This subset contains carefully curated examples where correct segmentation requires understanding the context of earlier interactions.
Conversations in MRSeg can extend up to $19$ rounds, though most interactions are shorter. 
Each type of conversation follows specific patterns, \textit{i}.\textit{e}., hierarchical conversations progress from whole objects to parts, positional relationships focus on spatial arrangements, and interactional relationships capture dynamic relationships between objects.

\paragraph{M\textsuperscript{4}Seg}
The M\textsuperscript{4}Seg \cite{prima} contains 224,393 samples, where the images are sourced from ADE20K-Part-234 \cite{wei2024ov,ade20k}, Pascal-Part \cite{chen2014detect}, and PACO-LVIS \cite{ramanathan2023paco}. 
This dataset aims to benchmark the multi-image RS capabilities, where each sample contains a multi-image set and one implicit text query.
To obtain similar images to construct multi-image sets, cosine similarity was computed on DINOv2 \cite{dinov2} features to identify visually related images. 
Afterward, nearest-neighbor sampling and object category-based sampling were utilized to form multi-image sets that share similar semantics.
Based on these multi-image sets, query-answer pairs were generated by GPT-4o, prompted with a complete list of all the objects and parts in the multi-image sets. 
To alleviate hallucination during query-answer pair generation, M\textsuperscript{4}Seg implemented a rule-based filtering that eliminated 38,873 question-answer pairs (14.74\% of the initial collection) that exhibited issues such as missing groundings, incorrect object references, or incomplete image coverage.
On average, each query-answer pair involves $5.83$ targets, with individual pairs containing up to $16$ targets.

\paragraph{MUSE}
The Multi-target Understanding and Segmentation Evaluation (MUSE) dataset \cite{pixellm} aims to evaluate the multi-target image RS capabilities, with instance-level annotations and detailed object descriptions directly linked to segmentation masks.
Formally, MUSE contains 246,000 query-answer pairs that cover 910,000 instance segmentation masks from the LVIS dataset \cite{lvis}, with an average of 3.7 targets per answer. 
The dataset is divided into training (239,000), validation (2,800), and testing (4,300) splits, with the test set further classified by query complexity (fewer or more than three targets). 
MUSE was constructed through GPT-4V-aided curation, where GPT-4V analyzes images with pre-existing mask annotations from LVIS \cite{lvis}, selecting instances to construct natural and contextually relevant query-answer pairs. 
Then, MUSE undergoes quality control through both automated LLM-based filtering and human verification to ensure that query-answer pairs follow intuitive reasoning patterns.

\paragraph{GranD and GranD\textsubscript{f}}
The Grounding-Anything Dataset (GranD) and its fine-tuning counterpart GranD\textsubscript{f} \cite{glamm} aims to evaluate and train RS models to generate natural language responses together with the corresponding segmentation masks across multiple targets. 
Created through a four-level annotation pipeline, GranD encompasses 11 million images with 810 million segmented regions and 7.5 million unique concepts from the SA-1B dataset \cite{sam1}. 
The hierarchical annotation process begins with object localization and attributes (Level-1), progresses to relationships and landmarks (Level-2), builds hierarchical scene graphs for dense captioning (Level-3), and concludes with contextual insights (Level-4). 
For fine-tuning purposes, GranD\textsubscript{f} comprises 214,000 image-grounded text pairs derived from existing datasets (RefCOCOg \cite{refcocog}, PSG \cite{psg}, and Flickr30K \cite{flickr30k}) enhanced with GPT-4-generated annotations, plus 1,000 high-quality manual annotations. 
GranD\textsubscript{f} contains 2,500 validation and 5,000 test samples. 

\paragraph{FP-RefCOCO(+/g)}
The FP-RefCOCO, FP-RefCOCO+, and FP-RefCOCOg datasets \cite{sesame} can evaluate RS models' capability to handle false premises where the queried object does not exist in the image. 
They extend the original RefCOCO \cite{refcoco}, RefCOCO+ \cite{refcoco}, and RefCOCOg \cite{refcocog} benchmarks by augmenting them with contextually relevant false premise queries. 
Rather than using naive random sampling for negative examples, GPT-3.5-turbo is used in the dataset construction stage to generate false premise queries by modifying a single element (object, adjective, or relation) in positive referring expressions, maintaining contextual relevance. 
The resulting datasets maintain nearly 1:1 positive/negative sample ratios with the same train/test/val splits as their original counterparts, comprising 276K sentences for FP-RefCOCO, 277K for FP-RefCOCO+, and 186K for FP-RefCOCOg. 

\paragraph{PRIST}
PRIST (Pixel-level Reasoning Segmentation dataset based on Multi-Turn conversation) \cite{miras} introduces a benchmark dataset focused on fine-grained image RS segmentation through progressive multi-turn dialogues.
Unlike other image RS benchmarks that primarily target region-level segmentation with single-turn reasoning, PRIST enables the evaluation of evolving user intent across conversation turns, containing 24k high-quality utterances, 8.3k multi-turn conversational scenarios, and corresponding pixel-level segmentation targets. 
The dataset was constructed through a three-step automated annotation pipeline that extracts visual elements, builds hierarchical reasoning trees, and generates contextually relevant dialogues. 
PRIST images are sourced from TextCaps \cite{sidorov2020textcaps} and annotated with fine-grained segmentation masks, with 53\% of the targets fine-grained (compared to 41\% small targets in COCO). 
Each conversation contains an average of 4 turns, with a maximum of 7, and features an average text length of 477 tokens. 
This structure enables the PRIST dataset to simulate real-world interactions while providing coherent reasoning chains for pixel-level target localization, establishing a valuable benchmark for evaluating both segmentation precision and reasoning capabilities in conversational contexts.

\paragraph{MMR}
The multi-target and multi-granularity reasoning (MMR) dataset \cite{jang2025mmr} addresses the limitations of existing image RS benchmark datasets that focus primarily on single-target or object-level reasoning by supporting both multiple target objects and fine-grained part-level reasoning.
Specifically, MMR comprises 194K complex and implicit queries in 57,643 images with the corresponding masks sourced from PACO-LVIS \cite{ramanathan2023paco}. 
The dataset was constructed through a GPT-4V-assisted generation pipeline that creates intricate question-answer pairs that require reasoning about both objects and their constituent parts. 
MMR includes a total of 75 object categories and 445 part categories, with each question-answer pair containing an average of 1.8 targets. 
This hierarchical generation approach enables models to reason about objects and their detailed parts within a single text query, facilitating more natural and context-aware interactions. 
The dataset is divided into training sets (154,127 pairs), validation sets (8,194 pairs), and test sets (32,077 pairs), with the test set further categorized into object-only, part-only, and mixed subsets to provide a benchmark for evaluating multi-granularity reasoning capabilities.

\section{Video Reasoning Segmentation}
\subsection{Task Definition}

Video RS extends image RS to the temporal dimension, aiming to segment and track objects in videos based on implicit text queries that require complex reasoning capabilities. 
Given an input video $\mathbf{X}_\text{video} = \{\mathbf{x}_t\}_{t=1}^T \in \mathbb{R}^{T \times H \times W \times 3}$ of $T$ frames, where each frame $\mathbf{x}_t$ has a dimension of $H \times W \times 3$, and an implicit text query $Q$.
RS aims to produce a binary mask sequence $\mathbf{M} = \{\mathbf{m}_t\}_{t=1}^T \in \mathbb{R}^{T \times H \times W}$ representing the target object across all frames with each $\mathbf{m}_t$ follows the definition in Eq.~\eqref{eq:image_rs_mask}.
Overall, the video RS can be formulated as:
\begin{equation}
\mathbf{M} = \varphi_{\theta}(\mathbf{X}_\text{video},Q),
\end{equation}
where $\varphi_{\theta}(\cdot)$ denotes the RS model with parameters $\theta$.
The task shares a similar formulation with referring video object segmentation \cite{ding2023mevis,seo2020urvos}. 
But, unlike direct text queries in referring segmentation that rely on explicit visual attributes, video RS necessitates reasoning to interpret implicit text queries. 
For instance, queries like ``\textit{the car powered by electricity}'' require an understanding of world knowledge beyond basic visual recognition, while queries like ``\textit{which car is most likely to win the race}'' demand reasoning about temporal dynamics and future predictions in the video context.
Moreover, the pixel-level segmentation must maintain temporal consistency while satisfying the reasoning requirements across frames. 
This introduces additional complexity compared to image RS, as the model must therefore integrate temporal understanding with complex reasoning to generate coherent mask sequences that accurately track the target objects throughout the video duration.

\subsection{Methods}
Although all the image RS can be directly applied to video RS tasks by processing each frame separately, it can lead to heavy computational costs because of the reliance on MLLMs and the results' lack of spatial consistency. 

\paragraph{TrackGPT}
TrackGPT \cite{trackgpt} is an online video RS approach that processes each frame sequentially.
At its core, TrackGPT employs a fine-tuned MLLM to comprehend implicit text queries and reason about target objects by expanding the MLLM's vocabulary with two specialized tokens, namely \texttt{<TK>} for referring queries and \texttt{<PO>} for tracking evaluation. 
For each video, the MLLM processes the first frame along with the implicit text query to generate these tokens, which are then projected into embeddings that guide the mask decoder in generating segmentation masks.
In response to environmental changes, a rethinking mechanism continuously monitors the quality tracking through a pretend score generated alongside the segmentation mask from the \texttt{<PO>} by the mask decoder.
When this score indicates deviation from the query's intent, it triggers the MLLM to reassess the current frame and update its tracking strategy. 
Complementing this, cross-frame propagation preserves temporal consistency while adapting to appearance changes by propagating and updating referring embeddings between consecutive frames.

\begin{figure}[htbp!]
    \centering
    \includegraphics[width=1.0\linewidth]{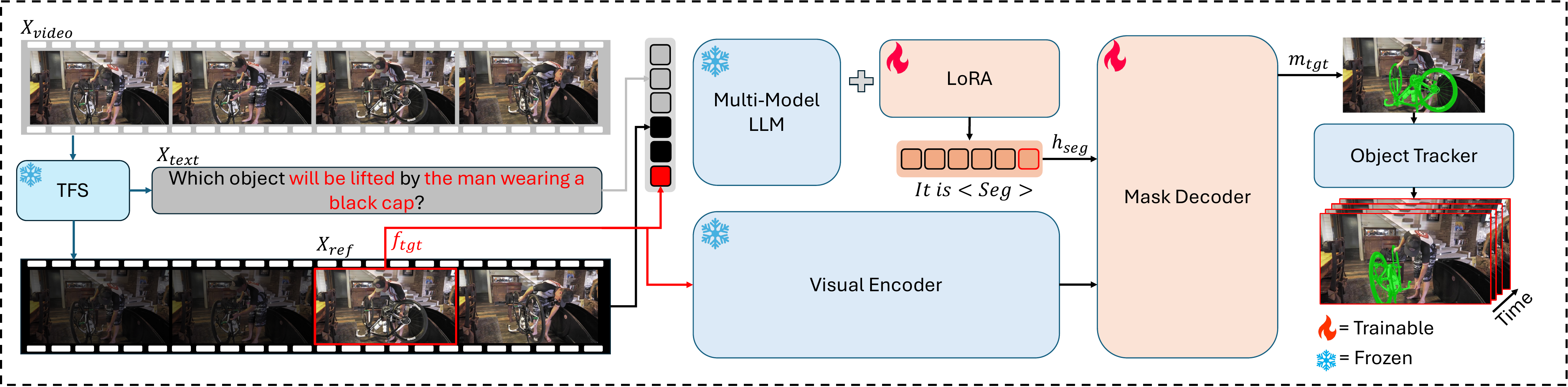}
    \caption{
    Overview of the VISA \cite{visa}. 
    }\label{fig:visa}
\end{figure}

\paragraph{VISA}
As shown in Fig.~\ref{fig:visa}, VISA formulates the video RS as image RS on the target frame and then performs temporal propagation to other frames \cite{visa}.
Unlike previous approaches that process single frames sequentially (\textit{e}.\textit{g}., TrackGPT \cite{trackgpt}) or employ heavy temporal pooling (\textit{e}.\textit{g}., Video-ChatGPT \cite{maaz2023video}), VISA is an offline method that introduces a text-guided frame (TFS) sampler to select the most relevant frame to perform the image RS.
Specifically, TFS leverages a frozen LLaMA-VID \cite{llamavid} to process the entire video sequence $\mathbf{X}_\text{video}$ and identify only one frame where the target objects are most distinguishable to perform image RS, \textit{i}.\textit{e}. the target frame. 
In addition to the target frame, multiple reference frames are selected, where target objects are also identified, to help the image RS in the target frame by providing a long-term temporal correspondence. 
Given the implicit text query, target frame, and reference frames, a fine-tuned MLLM generates text output containing a \texttt{<SEG>} token \cite{lisa}, whose embedding is decoded by a SAM-based mask decoder to RS mask for the target frame. 
The masks are then propagated bi-directionally to all remaining frames using an object tracker \cite{cheng2022xmem}, enabling consistent segmentation throughout the video sequence. 
This architecture effectively balances computational efficiency with segmentation quality by focusing the model's attention on the most relevant temporal context while maintaining sufficient spatial detail for accurate mask generation. 

\paragraph{ViLLa}
ViLLa \cite{villa} is an online end-to-end video RS method that can segment multiple instances of the same or different semantic categories. 
ViLLa comprises two innovations: a context aggregation module and a video frame decoder. 
The context aggregation module enhances the implicit text query's embeddings by incorporating relevant visual features from the current frame, creating a rich representation before processing by the MLLM. 
To capture temporal dynamics effectively, ViLLa implements multi-scale segmentation tokens that operate at both frame and video levels, enabling modeling of local and global temporal dependencies. 
These tokens are then refined through a video-frame decoder that facilitates bidirectional information exchange between frame-level and video-level representations using momentum-based aggregation. 
The decoder incorporates masked cross-attention, self-attention, and feedforward networks to progressively refine segmentation embeddings across the temporal dimension, ensuring robust adaptation to appearance changes and object motion throughout the video sequence.
The entire ViLLa is trained end-to-end using a combination of text generation and segmentation mask losses, enabling joint optimization of reasoning and segmentation capabilities. 

\paragraph{VideoLISA}
VideoLISA \cite{videolisa} is an offline end-to-end video RS method that aims to equip MLLM with the ability to understand the temporality of the video by sparse dense sampling and the ability to segment the data using the one-token seg-all approach. 
VideoLISA comprises a visual tokenizer initialized from LLaVA \cite{llava}, an MLLM, an image encoder initialized from SAM \cite{sam1}, and a promptable mask decoder. 
The sparse dense sampling is proposed to balance computational demands with video understanding by maintaining full-resolution features for a subset of dense frames while downsampling the remaining sparse frames, by leveraging the inherent temporal redundancy in videos. 
After sparse dense sampling, the number of tokens for the video is reduced while preserving enough spatial details and temporal dynamics.
Those tokens, together with the text tokens, are then concatenated and processed by MLLM.
The one-token-seg-all approach employs a specialized \texttt{<TRK>} token that enables unified segmentation across multiple frames, trained to capture both semantic information and temporal dynamics. 
During inference, MLLM generates this \texttt{<TRK>} token to guide the mask decoder in producing consistent segmentation masks across all frames. 
To improve mask quality, VideoLISA incorporates a post-processing with XMem++ \cite{xmem++} that refines the segmentation results using dense frames as a reference.

\paragraph{InstructSeg}
InstructSeg \cite{instructseg} unifies RS across both image and video domains through a single MLLM together with an object-aware video perceiver and a vision-guided multi-granularity text fusion module in an end-to-end manner. 
Initially, the CLIP \cite{clip} image encoder in the MLLM converts the vision input \textit{i}.\textit{e}., image or key frame of the video into vision tokens.
The object-aware video perceiver then extracts temporal and object-specific information from reference frames, which are the frames selected from the video to provide important contextual and temporal information to guide the RS \cite{visa} or the image itself.
Specifically, it uses learnable queries to condense the information from reference frames into fixed-length tokens, which are then processed alongside text tokens, learnable mask tokens, and vision tokens by the MLLM. 
To better handle complex implicit text queries, the vision-guided multi-granularity text fusion module integrates both global and detailed information from queries with fine-grained visual guidance, rather than simply averaging or summarizing all text token embeddings as done in previous approaches \cite{lisa,visa}. 
The segmentation decoder finally generates masks and scores based on each of the mask embeddings and multi-granularity text embeddings separately, where the final mask is selected by applying a threshold to the mask scores. 
InstructSeg is efficient as it preserves pre-trained knowledge by applying LoRA \cite{lora} to fine-tune only specific components while keeping visual encoders frozen.

\paragraph{VRS-HQ}
VRS-HQ focuses on improving temporal consistency and segmentation quality in video RS through an end-to-end framework \cite{vrshq}.
Unlike previous video RS methods \cite{visa,videolisa} that rely on external models for keyframe selection and mask propagation, VRS-HQ focuses on improving temporal coherence and segmentation quality by leveraging hierarchical token representations and dynamic temporal aggregation. 
Instead of using a single \texttt{<SEG>} token from MLLM for performing segmentation, VRS-HQ introduces a multi-level token design with multiple frame-level \texttt{<SEG>} tokens and one temporal-level \texttt{<TAK>} token to capture both local spatial details and global temporal dynamics. 
For token fusion, it proposes a temporal dynamic aggregation that merges frame-level embeddings from \texttt{<SEG>} tokens into the temporal-level embedding from \texttt{<TAK>} token, using cosine similarity-based weights to enable better handling of inter-frame motion and appearance changes. 
VRS-HQ also introduces token-driven keyframe selection to identify optimal keyframes, eliminating the need for external keyframe detection models such as LLaMA-VID \cite{llamavid} in VISA \cite{visa}.
To be more specific, it combines token similarity scores with SAM2-generated occlusion scores \cite{sam2} to indicate the confidence of the target's presence in the current frame. 
For mask generation, VRS-HQ integrates the fused temporal token with the image embedding of the keyframe from the SAM2 image encoder, and then propagates it into other frames in an end-to-end manner, where the fused temporal token guides cross-frame propagation through SAM2's memory mechanism \cite{sam2}. 

\paragraph{MoRA}
Motion-Grounded Video Reasoning Assistant (MoRA) \cite{mora} is a video RS framework that builds upon LISA's architecture \cite{lisa}.
Specifically, MoRA employs a frozen CLIP-based visual encoder to process video frames sequentially, followed by a spatio-temporal pooling of Video-ChatGPT \cite{maaz2023video} to efficiently aggregate temporal information. 
To handle the unique temporal aspects of video RS, MoRA introduces a specialized \texttt{<LOC>} token alongside LISA's \texttt{<SEG>} token. 
While \texttt{<SEG>} guides spatial mask generation, \texttt{<LOC>} enables temporal boundary localization through an additional MLP head that generates binary temporal masks to prevent false activations during frame-wise mask decoding. 
For efficient training, MoRA employs a two-stage process: first, pre-training on referring video segmentation datasets without temporal localization, then fine-tuning with the complete architecture, including temporal components on video RS data. 

\paragraph{ViCaS}
ViCaS \cite{vicas} unifies holistic video understanding with pixel-level localization through captioning with video RS. 
Unlike previous approaches that address these tasks separately, ViCaS simultaneously handles high-level video description and fine-grained object segmentation. 
The architecture of ViCaS consists of three components: (1) a multimodal LLM for processing both text and video, (2) a vision backbone that extracts slow-fast video features through temporal and spatial pooling, and (3) a segmentation network with a mask decoder. 
For processing video content, ViCaS employs a slow-fast pooling that balances temporal and spatial information by maintaining temporally dense but spatially condensed features alongside spatially dense but temporally sparse features. 
ViCaS ensures efficient token representation while preserving information across both dimensions. 
During inference, ViCaS can generate detailed video descriptions for the captioning task or output specialized \texttt{<SEG>} tokens for each target object described in the text query.
These tokens are then processed by the mask decoder alongside multi-scale features from the segmentation backbone to generate temporally consistent object masks. 

\paragraph{JiT}
Just-in-Time (JiT) Digital Twins \cite{jitdt} introduce an agent framework that disentangles perception and reasoning for online video RS without requiring LLM fine-tuning \cite{shen2025position}.
Unlike previous approaches that rely heavily on MLLMs for both visual perception and reasoning, this method employs an LLM planner to construct a directed acyclic graph (DAG) that determines which specialist vision models (such as SAM \cite{sam1} and DepthAnything \cite{depthanything}) to address understanding of query's semantic, spatial, and temporal requirements.
The core innovation is the dynamic scene graph structure of the digital twin \cite{shen2025position}, which preserves fine-grained semantic, spatial, and temporal information through specialized node attributes and edge relationships, rather than compressing visual content into fixed-length token sequences as in all previous RS methods.
To accomplish RS, JiT employs complementary LLM components that reformulate video RS as digital twin component retrieval.
Specifically, a base LLM handles semantic reasoning for high-level understanding, while an LLM-coder generates executable code for complex spatial and temporal reasoning operations that are applied directly to the digital twin to retrieve the corresponding component based on the implicit text query.

\begin{table*}[htbp]
\caption{
Comparison of evaluation metrics for video RS across multiple dimensions, including their primary focus, mathematical formulation, 
original purpose/domain, advantages, limitations, implementation complexity, and usage frequency in video RS literature.
}
\label{table:video_rs_metrics}
\centering
\resizebox{\linewidth}{!}{
\begin{tabular}{l|c|c|c|c|c|c|c}
\toprule
Metric & Primary Focus & Formula & Original Purpose/Domain & Advantages & Limitations & Implementation Complexity & Usage Frequency \\
\hline
\rowcolor{gray!5}
Region Similarity ($\mathcal{J}$) & Spatial Accuracy & $\mathcal{J} = \frac{|P \cap G|}{|P \cup G|}$ & 
Object tracking, video segmentation & 
Direct measure of overlap; intuitive interpretation & 
Insensitive to boundary precision; biased toward large objects & 
Low & 
Very Common \\
Boundary Accuracy ($\mathcal{F}$) & Contour Precision & $\mathcal{F} = \frac{2 \cdot \text{Precision} \cdot \text{Recall}}{\text{Precision} + \text{Recall}}$ & 
Video object segmentation & 
Captures fine-grained boundary details; shape-sensitive & 
Computationally intensive; less intuitive than IoU & 
Medium & 
Very Common \\
\rowcolor{gray!5}
$\mathcal{J}\&\mathcal{F}$ Score & Balanced Assessment & $\mathcal{J}\&\mathcal{F} = \frac{\mathcal{J} + \mathcal{F}}{2}$ & 
Video object segmentation & 
Combines region and boundary metrics; balanced evaluation & 
Equally weights potentially distinct aspects; masks individual metric failures & 
Low & 
Very Common \\
Recall Scores ($\mathcal{J}$-$\mathcal{R}$, $\mathcal{F}$-$\mathcal{R}$) & Retrieval Efficacy & $\mathcal{J}\text{-}\mathcal{R} = \frac{|\{i | \mathcal{J}_i > \tau_J\}|}{N}$ & 
Video object retrieval & 
Thresholded performance assessment; retrieval-focused & 
Threshold-dependent; binary success measure & 
Medium & 
Common \\
\rowcolor{gray!5}
Average Precision (AP) & Temporal Consistency & $\text{AP} = \frac{1}{N} \sum_{i=1}^{N} P(r_i)$ & 
Video instance segmentation & 
Comprehensive precision-recall assessment; confidence-aware & 
Complex implementation; requires confidence scores & 
High & 
Common \\
Average Recall (AR) & Completeness & $\text{AR} = \frac{1}{|\mathcal{T}|}\sum_{t \in \mathcal{T}} \frac{|TP_t|}{|TP_t| + |FN_t|}$ & 
Video instance segmentation & 
Multi-threshold retrieval assessment; completeness-focused & 
Less sensitive to false positives; requires threshold selection & 
High & 
Common \\
\rowcolor{gray!5}
Robustness Score ($R$) & Negative Sample Handling & $\text{R} = 1 - \frac{\sum_{M \in \mathcal{M}_{\text{neg}}}|M|}{\sum_{M \in \mathcal{M}_{\text{pos}}}|M|}$ & 
Robust video-language understanding & 
Quantifies false alarm resistance; handles negative queries & 
Requires positive samples for normalization; relative measure & 
High & 
Emerging \\
\bottomrule
\end{tabular}
}
\end{table*}

\subsection{Evaluation Metrics}

Table \ref{table:video_rs_metrics} provides an overview of the evaluation metrics used in video RS.

\paragraph{Region Similarity ($\mathcal{J}$)}
Region similarity, denoted as $\mathcal{J}$, measures the intersection over union between the predicted segmentation mask and the ground truth mask \textit{i}.\textit{e}.,
\begin{equation}
\mathcal{J} = \frac{|P \cap G|}{|P \cup G|},
\end{equation}
where $P$ is the predicted mask and $G$ is the ground truth mask. 
Region similarity mainly quantifies how accurately the predicted region overlaps with the target region, providing a direct assessment of spatial accuracy in reasoning-based segmentation. %
When extending from image-wise evaluation onto the multi-frame videos, the mean $\mathcal{J}$ across all frames is typically reported to measure temporal consistency, namely
\begin{equation} 
\mathcal{J}_{\text{mean}} = \frac{1}{T}\sum_{t=1}^{T} \mathcal{J}_t, 
\end{equation} 
where $T$ is the total number of frames and $\mathcal{J}_t$ is the region similarity with respect to the $t$-th frame.

\paragraph{Boundary Accuracy ($\mathcal{F}$)}
Boundary accuracy, denoted as $\mathcal{F}$, evaluates the performance of delineating the object boundaries in video RS tasks.
The $\mathcal{F}$ is computed by treating the boundary prediction as a binary classification on the object boundary pixels, namely,
\begin{equation} 
\mathcal{F} = \frac{2 \cdot \text{Precision} \cdot \text{Recall}}{\text{Precision} + \text{Recall}}, 
\end{equation} 
where precision and recall are calculated based on the distance between boundary pixels of the predicted mask and the ground truth mask. 
The boundary accuracy focuses specifically on the contour accuracy of the predicted mask compared to the ground truth, providing an understanding of how well RS models capture the exact shape of target objects when responding to the implicit query.

\paragraph{$\mathcal{J}$\&$\mathcal{F}$ Score}
The $\mathcal{J}$\&$\mathcal{F}$ score is the arithmetic mean of region similarity $\mathcal{J}$ and boundary accuracy $\mathcal{F}$, providing a balanced evaluation of both segmentation region overlap and boundary precision, formulated as
\begin{equation}
\mathcal{J}\&\mathcal{F} = \frac{\mathcal{J} + \mathcal{F}}{2}, 
\end{equation}
This metric offers an assessment of the quality of the segmentation, considering both the accuracy of the overall region and the precision of the boundary delineation. 
%
%
Higher $\mathcal{J}\&\mathcal{F}$ scores indicate superior performance in both aspects of segmentation quality within the context of instruction-based or reasoning-based tracking.

\paragraph{Recall Scores ($\mathcal{J}-\mathcal{R}$ and $\mathcal{F}-\mathcal{R}$)} %
Recall-based metrics $\mathcal{J}-\mathcal{R}$ and $\mathcal{F}-\mathcal{R}$ are used to evaluate the model's ability to retrieve correct target objects across all instances. %
$\mathcal{J}-\mathcal{R}$ represents the recall measurement for region similarity,
\begin{equation} 
\mathcal{J}-\mathcal{R} = \frac{|\{i | \mathcal{J}_i > \tau_J\}|}{N}, \end{equation} 
where $\tau_J$ is a threshold (typically 0.5), $N$ is the total number of samples, and $\mathcal{J}_i$ is the region similarity for the $i$-th sample. 
Similarly, $\mathcal{F}-\mathcal{R}$ indicates the recall for boundary accuracy, namely 
\begin{equation} 
\mathcal{F}-\mathcal{R} = \frac{|\{i | \text{F}_i > \tau_F\}|}{N}, \end{equation} 
where $\tau_F$ is a threshold for boundary accuracy. 
Higher recall scores indicate better reasoning abilities when interpreting abstract or implicit tracking instructions.

\paragraph{Average Precision (AP)}
Average Precision (AP) aims to assess both localization accuracy and classification correctness across temporal dimensions in video RS \cite{villa}.
Unlike previous IoU-based metrics that focus solely on spatial overlap, AP evaluates the quality of predicted segmentation masks throughout a video sequence by computing precision at different recall levels and averaging the results.
Formally, for a video with $T$ frames, AP is calculated as:
\begin{equation}
\text{AP} = \frac{1}{N} \sum_{i=1}^{N} P(r_i),
\end{equation}
where $N$ is the number of different recall thresholds, $r_i$ is the $i$-th recall threshold, and $P(r)$ is the maximum precision at the recall level $r$. 
%
%
The function $P(r)$ represents the highest precision value achievable at recall level $r$ or higher, capturing the precision-recall tradeoff at various operating points.
For video instance RS segmentation tasks, AP is computed by first matching predicted mask tracklets with ground truth tracklets using a bipartite matching algorithm that maximizes the IoU across frames:
\begin{equation}
\text{IoU}_{\text{video}} = \frac{\sum_{t=1}^{T} |P_t \cap G_t|}{\sum_{t=1}^{T} |P_t \cup G_t|},
\end{equation}
where $P_t$ and $G_t$ are the predicted and ground truth masks at frame $t$, respectively.
After this matching process, predictions are sorted by their confidence scores in descending order.
For each confidence threshold, true positives (predictions matched with $\text{IoU}_{\text{video}}$ above a predetermined threshold) and false positives are identified. 
At each threshold, the precision and recall values are computed, creating a precision-recall curve. 
AP is then calculated as the area under this precision-recall curve, effectively measuring the model's ability to balance precision and recall across the entire video while maintaining temporal consistency in both reasoning and segmentation.

\paragraph{Average Recall (AR)}
Average Recall (AR) complements AP by measuring the model's ability to retrieve all relevant target objects in a video RS task \cite{villa}.
AR is computed as the average recall across different IoU thresholds:
\begin{equation}
\text{AR} = \frac{1}{|\mathcal{T}|}\sum_{t \in \mathcal{T}} \frac{|TP_t|}{|TP_t| + |FN_t|},
\end{equation}
where $\mathcal{T}$ is the set of IoU thresholds, and $TP_t$ and $FN_t$ represent true positives and false negatives at the threshold $t$, respectively.
In video RS, high AR indicates that the model effectively captures all instances of the target objects described in implicit reasoning queries throughout the video sequence.

\paragraph{Robustness Score}
The Robustness Score ($R$) is designed for video RS tasks that measure a model's ability to correctly handle negative video language pairs \cite{li2023robust}. 
Unlike traditional video RS metrics that focus solely on positive pairs, $R$ specifically evaluates how effectively a model avoids false alarm segmentations when the object described in the implicit query does not exist in the video.
Mathematically, $R$ is defined as:
\begin{equation}
\text{R} = 1 - \frac{\sum_{M \in \mathcal{M}_{\text{neg}}}|M|}{\sum_{M \in \mathcal{M}_{\text{pos}}}|M|},
\end{equation}
where $\mathcal{M}_{\text{neg}}$ and $\mathcal{M}_{\text{pos}}$ represent the sets of segmented masks in negative and positive videos respectively, and $|M|$ denotes the foreground area of mask $M$.
The total foreground area of positive videos serves as a normalization term.
This metric assesses robustness rather than just using alignment accuracy, as it directly measures the degree of misclassified pixels in negative videos.
Ideally, a robust model should treat all negative videos as background (no foreground segmentation is required), resulting in an $R$ score of 1.0, indicating perfect discrimination between positive and negative video-language pairs.

\subsection{Datasets and Benchmarks}

\begin{figure}[htbp!]
    \centering
    \includegraphics[width=1.0\linewidth]{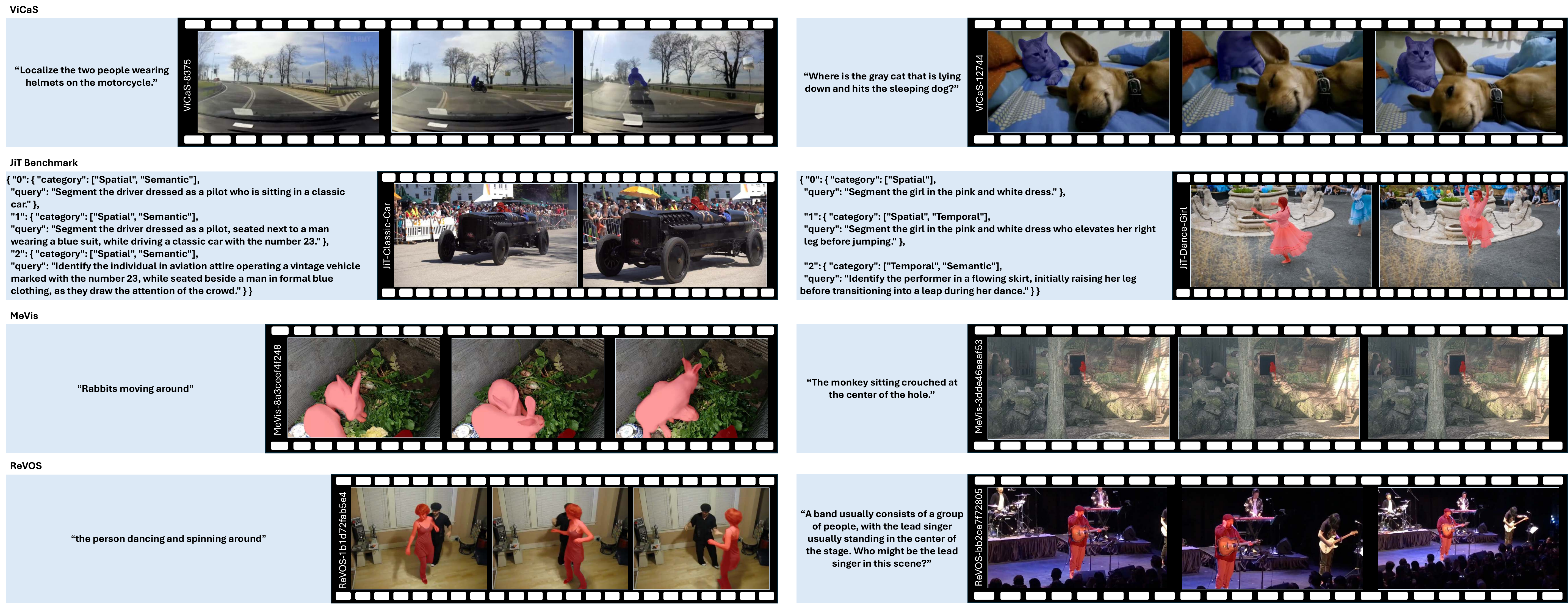}
    \caption{
    Illustrative example of video RS data.
    }\label{fig:benchmark2}
\end{figure}

Fig.~\ref{fig:benchmark2} shows representative samples of video RS data.

\paragraph{Refer-Youtube-VOS}
Refer-Youtube-VOS \cite{seo2020urvos} is a benchmark initially designed for referring video object segmentation, built upon Youtube-VOS \cite{xu2018youtube}. 
It contains 27,899 referring expressions for 7,451 objects in 3,975 high-resolution videos with 94 common object categories.
Each video spans approximately 3-6 seconds with pixel-level instance segmentation annotations at every 5 frames.
Refer-Youtube-VOS has two distinct annotation types, namely ``full video expressions'' where annotators watch the entire video, and ``first-frame expressions'' where only the first frame is viewed for annotation, resulting in different expression characteristics. 
Full-video expressions (averaging 10.0 words) can incorporate dynamic information, while first-frame expressions (averaging 7.5 words) focus primarily on appearance attributes.

\paragraph{Refer-DAVIS\textsubscript{17}}
Refer-DAVIS\textsubscript{17} \cite{khoreva2019video} extends DAVIS\textsubscript{17} \cite{pont20172017} by including the language descriptions of specific objects in each video.
Similarly to Refer-Youtube-VOS, Refer-DAVIS\textsubscript{17} was also collected under two different annotation settings, namely descriptions based only on the first frame and based on the full video. 
Two non-expert annotators provided first-frame descriptions, while a separate annotation collected full-video descriptions. 
On average, each video contains 7.5 referring expressions per target object, and each expression takes approximately 5 seconds for an annotator to provide. 
The first-frame and full-video expressions have average lengths of 5.5 and 6.3 words, respectively, with full-video descriptions containing more verbs and being generally more complex.

\paragraph{InsTrack}
InsTrack \cite{trackgpt} is a benchmark that involves providing tracking queries requiring the model to perform reasoning to identify and track the appropriate target. 
It contains 143 video sequences for instruction tuning and 40 for evaluation, with each video accompanied by 6 tracking instructions, resulting in over 1,000 instruction-video pairs. 
The videos are sourced from Refer-DAVIS\textsubscript{17} \cite{khoreva2019video} (32 sequences) and VIPSeg \cite{miao2022large} (151 sequences), with re-annotated implicit instructions for the objects of interest. 
Those instructions are more aligned with natural human queries (\textit{e}.\textit{g}., ``\textit{Who will win the race?}''), requiring models to reason about visual content before performing tracking.

\paragraph{MeViS}
MeViS \cite{ding2023mevis} is a benchmark for motion-expression guided video referring segmentation, where objects are segmented based on explicit descriptions of their movements rather than static attributes. 
Unlike previous video referring datasets that focus on salient objects with obvious static features, MeViS contains 2,006 videos with 8,171 objects and 28,570 motion expressions, resulting in 443k mask annotations.
Unlike InsTrack \cite{trackgpt}, MeViS focuses on longer videos (averaging 13.16 seconds) with multiple objects of similar appearance (4.28 objects per video), making it difficult to identify targets through saliency or category information alone.
The expressions in MeViS primarily describe motion attributes (\textit{e}.\textit{g}., ``\textit{The bird flying away}''), with validation criteria ensuring that targets cannot be identified from a single frame. 
Another unique feature of MeViS is its support for multi-object expressions, where one description can refer to multiple targets (averaging 1.59 objects per expression), thus making it more challenging than the InsTrack dataset.

\paragraph{ReVOS}
ReVOS \cite{visa} is the first benchmark for video RS that evaluates the reasoning abilities based on both video content and world knowledge.
It contains 35,074 pairs of object instructions in 1,042 different videos sourced from LV-VIS \cite{wang2024ov,wang2023towards}, MOSE \cite{ding2023mose}, OVIS \cite{qi2022occluded}, TAO \cite{dave2020tao}, and UVO \cite{wang2021unidentified}.
The ReVOS dataset is divided into training (626 videos) and validation (416 videos) sets, with text instructions falling into three categories: (1) 14,678 implicit queries requiring reasoning and inference, (2) 20,071 explicit queries for evaluating generalization to traditional referring tasks, and (3) 325 descriptions of non-existent objects for hallucination evaluation.
Unlike previous referring video segmentation datasets that use short phrases describing visual attributes, the queries of ReVOS are complex sentences, which therefore require understanding of video content, general knowledge of the world, and reasoning about temporal dynamics.

\paragraph{VideoReasonSeg}
VideoReasonSeg \cite{villa} is also a benchmark for video RS, comprising 1,934 video-instruction-mask tuples collected from diverse sources, including YouTube-VIS \cite{yang2019video}, OVIS \cite{qi2022occluded}, and LV-VIS \cite{wang2023towards}.
Each video is first annotated with implicit text instructions and high-quality target masks to test RS models' ability to reason across temporal dimensions.
The VideoReasonSeg dataset supports two evaluation paradigms: 1) multiple-choice QAs and 2) instruction-answer pairs with the corresponding segmentation masks.
For the multiple-choice format, each video is associated with two pairs of question-answer questions, where the questions require reasoning about the video content, and the options are carefully balanced in length to prevent the leakage of the answers.
The VideoReasonSeg dataset was constructed through a GPT-4V-aided automatic generation pipeline, where the LLM analyzed videos with their instance annotations to autonomously create question-answer pairs requiring reasoning about visual and temporal relationships.
Unlike static image reasoning datasets, VideoReasonSeg emphasizes temporal reasoning with questions that often involve motion information (\textit{e.g.}, ``\textit{a person wearing a white shirt riding a wave}''), requiring models to track objects while reasoning about their relationships and interactions.
The VideoReasonSeg dataset is divided into training (1,000 videos), validation (400 videos), and testing (534 videos) splits to facilitate RS model fine-tuning and benchmarking.

\paragraph{ReasonVOS}
As a benchmark for Video RS, ReasonVOS \cite{videolisa} contains 458 video-instruction-mask data samples with mask annotations sourced from existing datasets, including MOSE \cite{ding2023mose}, MeViS \cite{ding2023mevis}, VIPSeg \cite{miao2022large}, and BURST \cite{athar2023burst}. 
The annotation process involved first manually creating 105 samples as initial seed data, then using an LLM to rephrase and augment these explicit text queries, followed by human verification. 
Each language expression in ReasonVOS must encompass at least one of three aspects: (1) complex reasoning, (2) application of world knowledge, or (3) understanding of temporal dynamics. 
The ReasonVOS dataset consists of 91 videos with 205 short queries (typically attributive clauses describing specific objects) and 253 long queries (full sentences requiring reasoning). 
Although InsTrack \cite{instructseg} also involves reasoning in their queries, ReasonVOS provides a more comprehensive evaluation with a greater emphasis on the integration of world knowledge. 
Unlike ReVOS \cite{visa}, and VideoReasonSeg \cite{villa}, which also include subsets for model fine-tuning, ReasonVOS is specifically intended for zero-shot evaluation.

\paragraph{GroundMORE}
GroundMORE (Grounding via Motion Reasoning) \cite{mora} is a benchmark designed for motion-grounded video RS.
GroundMORE differs from other video RS datasets such as ReVOS \cite{visa} and VideoReasonSeg \cite{villa} by incorporating more motion context analysis, requiring reasoning about temporal boundaries, and abstract motion-related attributes.
The GroundMORE dataset contains 1,715 video clips with 7,577 questions, 249K object masks involving 3,942 different objects, and an average clip length of 9.61 seconds.
The videos are sourced from Youtube, spanning four diverse domains, namely family scenes (35.1\%), ball games (32.7\%), animal interactions (25.4\%), and outdoor activities (6.8\%).
Regarding the dataset annotation,  annotators first create motion-related expressions describing interactions in the videos.
Second, these expressions are converted into four distinct question types using GPT-4o: 1) causal questions explore motivations behind motions, 2) sequential questions probe the order of temporally adjacent motions, 3) counterfactual questions focus on hypothetical scenarios, and 4) descriptive questions address general scene understanding or abstract motion-related attributes.

\paragraph{ViCaS}
ViCaS \cite{vicas} contains 7,331 videos sourced from the Oops dataset \cite{epstein2020oops}, with each video annotated with detailed human-written captions (averaging 42.5 words) in which phrases referring to salient objects are grounded with temporally consistent segmentation masks. 
Professional annotators first write detailed captions and mark object phrases using a specific syntax; second, different annotators draw segmentation masks for each marked object. 
To balance annotation quality and cost, masks are drawn at 1 frame-per-second and then propagated to 30 FPS using SAM2 \cite{sam2}. 
The dataset contains 20,699 object tracks with 5.29M object masks and encompasses 4,411 unique nouns/noun-phrases. 
Unlike previous video RS datasets that focus on either high-level understanding or pixel-level segmentation, ViCaS supports evaluation of both capabilities through two benchmark tasks: (1) Video Captioning, which requires generating detailed descriptions of video content, and (2) Video RS, which involves predicting segmentation masks for multiple objects based on implicit text prompts.

\paragraph{JiT Benchmark}
The JiT benchmark \cite{jitdt} is designed to video RS across varying levels of reasoning complexity. 
It comprises 200 videos with 895 implicit text queries that span three reasoning categories, namely semantic, spatial, and temporal. 
Each query is further categorized into one of three progressive difficulty levels: Level 1 focuses on basic reasoning, Level 2 introduces two-step reasoning, and Level 3 requires complex multi-step reasoning with more than three inference steps. 
The dataset is based on videos and the corresponding masklets from DAVIS \cite{pont20172017} and SA-V test datasets \cite{sam2}, where each sample contains the source video sequence, ground-truth binary segmentation masks, implicit text query, reasoning category, and difficulty level. 
JiT benchmark is deliberately constructed to evaluate the model's capability to handle online processing with complex chains of reasoning by maintaining a balanced distribution across different reasoning types and difficulty levels. 

\section{Application of Reasoning Segmentation}

\subsection{Existing Applications}

\paragraph{Safety Monitoring}
RS enables enhanced safety monitoring by interpreting complex visual scenes through implicit text queries to identify security risks, safety violations, and regulatory non-compliance.
Previous work \cite{luo2025ground} demonstrated this ability in verifying scaffold safety compliance at construction sites. 
Specifically, RS is adapted to identify abstract scaffolding structure concepts, such as ``\textit{horizontal bottom tubes}'', ``\textit{cross bracings}'', and ``\textit{horizontal transverse tubes}'' by text queries.
The method improves LISA \cite{lisa} by combining shallow and deep image features to better distinguish similar tubular structures.

\paragraph{Underwater Images Segmentation}
Underwater environments present visual challenges including light attenuation, color shifts, and reduced visibility. 
Previous work \cite{yao2025language} applied LISA \cite{lisa} to underwater image segmentation, enabling the distinction between visually similar marine objects through text queries. 
They constructed the RefUDS dataset, pairing underwater images with detailed implict text query and corresponding segmentation masks for LISA fine-tuning on underwater conditions.

\paragraph{Operating Room Workflow Analysis}
RS facilitates automated analysis of operating room workflows through implicit query interpretation. 
Previous work \cite{shen2025operating} introduced ORDiRS (Operating Room Digital twin Representation for reasoning Segmentation), which leverages JiT \cite{jitdt} to analyze workflow bottlenecks. 
It can process queries such as ``\textit{segment non-essential personnel entering sterile fields}'' by identifying relevant staff, evaluating their positions relative to sterile zones, and generating appropriate segmentation masks.

\paragraph{Earth Observation}
Remote sensing applications benefit from the capabilities of RS in interpreting complex spatial relationships and applying domain knowledge.
EarthReason \cite{li2025segearth} is a benchmark that contains 5,434 annotated remote sensing images with more than 30,000 implicit question-answer pairs in 28 scene categories and resolutions from 0.5 to 153m. 
Their SegEarth-R1 integrates a hierarchical visual encoder for scale variation handling, an LLM for instruction interpretation, and a specialized mask generator for spatial correlation analysis. This enables complex reasoning tasks, such as identifying landslide-prone regions by simultaneously considering slope characteristics, vegetation coverage, and infrastructure proximity.

\subsection{Potential Applications}

While the previous subsection outlined existing applications, this section explores promising future directions for reasoning segmentation.

\paragraph{Video Forensics}
Traditional video forensics relies on explicit instructions and predefined categories, limiting effectiveness when investigating unconventional or contextually dependent targets. 
RS addresses these limitations by interpreting implicit queries and applying complex reasoning to visual evidence. 
Investigators can process queries such as ``\textit{Find bags unattended for longer than 15 minutes}'' to assess security protocols and identify suspicious activities. 
This capability reduces the manual analysis effort and enables more comprehensive forensic investigations under varying environmental conditions.

\paragraph{Video Anomaly Detection}
Current anomaly detection methods using MLLMs can identify and describe anomalies at the frame level \cite{lv2024videoanomalydetectionexplanation} but cannot precisely localize anomalous objects within frames using implicit queries \cite{yang2024follow}. 
RS addresses this limitation by processing queries such as ``\textit{Is there any anomaly?}'' while leveraging world knowledge about complex object relationships, as demonstrated in AnomalyRuler \cite{ye2024vera}. 
By providing both segmentation masks and textual explanations, RS enables precise localization of anomalous objects and supports complex reasoning about temporal anomalies that manifest through object interactions across multiple frames.

\paragraph{Smart Cities and Urban Planning}
Building on earth observation applications \cite{li2025segearth}, RS shows potential for smart city development. 
Urban planners can pose complex queries such as ``\textit{identify areas with insufficient green space relative to population density}'' or ``\textit{segment regions most vulnerable to urban heat island effects}.'' 
These queries require domain knowledge beyond basic object recognition to identify relationships between physical structures, human activities, and environmental factors. 
This RS capability supports more informed urban development strategies by enabling the complete analysis of complex urban systems.

\paragraph{Autonomous Navigation}
Autonomous vehicles and robots could benefit from RS capabilities when navigate complex environments \cite{gao2024survey}.
Rather than relying solely on predefined object categories, these systems could process queries like ``\textit{segment the safest path considering current weather conditions}'' or ``\textit{identify areas where pedestrians could unexpectedly cross}.'' 
This form of reasoning about dynamic environmental factors, causal relationships, and potential future states represents an advance beyond current perception methods that focus primarily on object detection and classification. 
By integrating temporal reasoning with spatial understanding, video RS could enable more adaptive and contextually aware navigation in unpredictable environments.
This would be particularly valuable in scenarios where standard recognition models might fail due to unusual conditions or circumstances that require human-like reasoning about safety, priority, and context-dependent behavior.

\section{Discussion}
\subsection{Limitations of Existing Works}

\paragraph{Task Settings}
Current RS formulations suffer from the following limitations in their scope. 
First, most existing RS methods focus primarily on single-step reasoning, with limited consideration for complex multi-step reasoning that more accurately reflect human cognitive processes \cite{cores,segzero}. 
Although some previous attempts, like CoReS \cite{cores} and Seg-Zero \cite{segzero}, have initiated exploration into multi-step reasoning, other RS approaches still operate within relatively constrained reasoning steps.
Specifically, deeper multi-step reasoning requires the decomposition of complex implicit user queries into a long chain of interconnected logical steps, with each step building upon previous inferences and contributing to subsequent reasoning stages. 
For example, when asked to segment ``\textit{the kitchen appliance that would be most energy efficient to heat a small meal},'' the RS model would sequentially: (1) identify all kitchen appliances in the image, (2) filter to those capable of heating food, (3) analyze their relative energy efficiency based on size and technology, (4) consider the context of a ``small meal,'' and finally (5) determine the optimal appliance. 
However, current RS methods typically compress these steps or handle them implicitly, limiting their transparency of reasoning.
Furthermore, current RS settings often overlook the inherent uncertainty in visual reasoning. 
Real-world visual scenes frequently contain ambiguities that require probabilistic reasoning or multiple valid interpretations. 
However, existing RS formulations treat reasoning as a deterministic process with a single correct answer, limiting their robustness in handling ambiguous queries or scenarios with multiple valid interpretations.
Furthermore, the current RS focuses on object-level reasoning, with limited exploration of fine-grained part-level reasoning or segmentation across different granularities simultaneously, which is particularly evident in video RS, as the temporal dynamics of objects and their parts introduce additional complexity.
Finally, there is inadequate consideration of the interactive nature of human reasoning in the current RS task formulation. 
Reasoning in real-world practice can involve iterative refinement based on human feedback. 
Although initial attempts like LISA++ \cite{lisa++} and MRSeg \cite{segllm} have made progress toward conversational interactions of RS, current approaches typically just accept a query and produce a segmentation mask. Bidirectional exchanges between users and RS models throughout the reasoning process remain largely unexplored.
Therefore, future directions can explore: (1) clarification in which the RS model can request additional information when queries are ambiguous (\textit{e}.\textit{g}., ``\textit{Do you mean the refrigerator that uses less electricity annually or the one that uses less energy per cooling cycle?}''), 
(2) explanations where the RS model articulates its reasoning steps and confidence levels, 
(3) refinement capabilities that allow users to correct intermediate reasoning steps (\textit{e}.\textit{g}., ``\textit{No, please consider only appliances on the counter}''), 
and (4) collaborative exploration of alternative reasoning paths.

\paragraph{Methods}
Existing RS methods are heavily based on external pre-trained segmentation models, such as the SAM encoder \cite{sam1,sam2}, leading to maintenance dependencies and potential bottlenecks in end-to-end training.
As these external models were not specifically designed for reasoning, they potentially create a mismatch between reasoning and segmentation capabilities. 
In addition, most RS methods rely on token-based approaches (\textit{e}.\textit{g}., <SEG>) to bridge reasoning and segmentation. 
Such ``embedding-as-mask'' paradigm introduced by LISA \cite{lisa} has become dominant, with limited exploration of alternative paradigms, which therefore restricts the diversity of RS approaches and potentially overlooks more effective solutions.
More direct integration between reasoning and segmentation components that can map reasoning outputs directly to spatial locations or architectures that jointly optimize for both tasks, rather than treating them as separate processes, shows to be an interesting future direction.
Moreover, the computational efficiency of current RS methods presents another limitation due to their heavy reliance on LLMs or MLLMs for both perception and reasoning. 
Specifically, video RS models require processing of video frames or employ heavy temporal pooling, making them computationally expensive and subsequently impractical for real-time applications. 
Even in image RS, the computational requirements of MLLMs are also challenging for deployment in resource-constrained environments such as end devices.
Furthermore, maintaining MLLM capabilities during fine-tuning for RS tasks is also another issue. 
Methods like LISA \cite{lisa} and its variants often experience a degradation in their original capabilities and reasoning performance when adapted for RS, caused by catastrophic forgetting. 
Although approaches like LLaVASeg \cite{llavaseg}, SAM4MLLM \cite{sam4mllm} and JiT \cite{jitdt} have attempted to address this by disentangling the reasoning from segmentation, these methods often introduce additional complexity in the inference stage.

\paragraph{Evaluation Metrics}
Existing RS evaluation metrics demonstrate an overemphasis on segmentation performance compared to reasoning quality. 
These metrics, such as cIoU, gIoU, and mIoU, focus primarily on the spatial correctness of segmentation masks rather than the correctness of the intermediate reasoning process that led to those masks.
Additionally, there is a lack of metrics to evaluate multi-step reasoning processes and their accuracy. 
Although some recent metrics, such as the GPT score \cite{popen,huang2024opera}, attempt to assess the quality of reasoning from its text response, these attempts often rely on subjective evaluations using LLM, introducing potential biases and inconsistencies, and also do not consider the intermediate reasoning steps.
For video RS specifically, current metrics inadequately capture the temporal consistency challenges specific to reasoning across frames as they do not address how well reasoning is maintained throughout a video sequence, especially when reasoning requires tracking causal or temporal relationships.
Furthermore, these evaluation metrics do not differentiate the types of reasoning required in different text queries. 
For example, an RS method might perform well in certain types of reasoning (\textit{e}.\textit{g}., spatial reasoning) while failing on others (\textit{e}.\textit{g}., counterfactual reasoning).
However, these RS metrics typically aggregate performance across all types of reasoning.
Lastly, there is insufficient attention to the evaluation of robustness against adversarial inputs or edge cases.

\paragraph{Datasets and Benchmarks}
Current datasets and benchmarks for RS relies on LLMs for the automated generation of reasoning queries and the corresponding ground-truth text responses such as ReasonSeg-Inst \cite{lisa++}, LLM-Seg40K \cite{llmseg}, MRSeg \cite{segllm}, and M\textsuperscript{4}Seg \cite{prima} which employ LLMs such as GPT-4 series to generate implicit text queries.
This method can potentially introduce the biases of LLM, as LLM-generated queries often reflect restricted reasoning patterns and stylistic uniformity that may not adequately represent the diversity and unpredictability of human-generated questions \cite{shen2025rvtbench,shen2025position}. 
Consequently, RS models trained on such datasets might excel at processing queries that match LLM generation patterns while struggling with real-world practice.
Furthermore, existing benchmarks focus on a narrow subset of reasoning categories. 
Most datasets concentrate on spatial reasoning (\textit{e}.\textit{g}., object relationships), attribute-based reasoning (\textit{e}.\textit{g}., color, size), and simple functional reasoning, while substantially under-representing reasoning types such as causal reasoning, counterfactual reasoning, abstract concept reasoning, analogical reasoning, quantitative reasoning and \textit{etc}. 
For example, queries that require an understanding of causality (\textit{e}.\textit{g}., ``\textit{segment the object that caused the spill}'') or hypothetical scenarios (\textit{e}.\textit{g}., ``\textit{segment the component that would malfunction first if this machine overheated}'') remain inadequately covered in existing RS benchmarks. 
This imbalance creates a gap between the reasoning capabilities required for real-world applications and the limited reasoning scopes evaluated in current benchmarks.
Finally, additional limitations include insufficient scale for many datasets, particularly for fine-tuning-based methods, with the original ReasonSeg \cite{lisa} containing only 1,238 samples, inadequate human verification of automatically generated annotations, and lack of standardization in benchmark construction and evaluations.

\subsection{Future Directions}

In light of the limitations discussed for current RS methods and benchmarks, we move on to the discussion of promising future research directions.

\paragraph{Expand Reasoning Capabilities}
Future research can look into deeper multi-step reasoning that explicitly decomposes complex text queries into interconnected logical steps to enable RS models for deeper cognitive processes, such as counterfactual reasoning, analogical reasoning, and causal inference. 
Furthermore, interactive RS frameworks would allow for more natural human-AI interaction during the reasoning process to potentially improve the user trust for the final segmentation masks. 
Specifically, developing clarification methods where RS models can request additional information for ambiguous queries and explanation capabilities where RS models articulate their reasoning steps would create more transparent and trustworthy interactions. 
Additionally, incorporating uncertainty estimation into reasoning would also enable RS models to handle ambiguous visual scenes and queries by providing confidence scores or multiple possible interpretations rather than forcing deterministic output. 
Finally, exploring cross-granularity reasoning that can transition between object-level and part-level understanding would enhance the flexibility and applicability of RS models, particularly for specialized domain applications such as medical image analysis and industrial inspection.

\paragraph{Evaluation Metrics}
Future work can also explore RS evaluation metrics that assess both segmentation quality and intermediate reasoning correctness. 
Specialized metrics for different reasoning categories (such as spatial, causal, counterfactual, \textit{etc}) would also enable better evaluation and highlight RS model strengths and weaknesses across various reasoning categories. 
For video RS, developing metrics that specifically evaluate the consistency of temporal reasoning across image frames would better capture the challenges of this modality.

\paragraph{Domain Specific Applications}
To bridge the gap between general RS problem settings and practical applications, future work can also focus on specialized RS approaches for domain-specific applications. 
For example, in medical image analysis, this could involve designing RS models capable of understanding complex anatomical relationships and functional interpretations to assist in diagnosis and treatment planning while improving interactions with RS clinicians. 
For autonomous navigation, designing RS methods that can reason about dynamic environments and predict potential hazards would help the safety and reliability analysis.

\paragraph{Integration with Other Modalities}
Extending RS beyond the visual and textual modalities is another direction, which can explore the integration of additional sensory inputs such as audio, depth information, thermal imaging, or other sensor data to enable better understanding and reasoning of the scene. 
For example, in autonomous driving scenarios, incorporating LiDAR data along with video data could improve reasoning about distances and safety considerations. 
In healthcare applications, the integration of patient medical records with visual data could enable more informed reasoning.

\section{Conclusion}
This survey has reviewed reasoning segmentation (RS), a new segmentation task that bridges the gap between traditional segmentation tasks and human-like reasoning by enabling models to identify the object to segment based on implicit text queries. 
Our analysis of 26 state-of-the-art RS methods across image and video reveals rapid progress in RS model designs, with approaches evolving from simple token-based designs to agents supporting multi-step reasoning and conversational capabilities. 
Current RS applications show promising results in diverse domains, including security monitoring, underwater image analysis, operating room workflow assessment, and earth observation, with potential for expansion to physical security, video anomaly detection, urban planning, autonomous navigation, and more. 
Future research can focus on designing standardized evaluation frameworks that assess both segmentation quality and intermediate reasoning correctness, domain-specific applications of RS, and integration with additional sensory modalities to allow for more complete scene understanding and reasoning.

\subsection*{Acknowledgments}
This work was supported in part by the JHU Amazon Initiative for Artificial Intelligence (AI2AI) fellowship program.

\bibliographystyle{model2-names.bst}
\bibliography{ref.bib}

\end{document}